\documentclass[]{interact}

\usepackage{epstopdf}% To incorporate .eps illustrations using PDFLaTeX, etc.
\usepackage{subcaption}

\usepackage{hyperref}
\usepackage{natbib}% Citation support using natbib.sty
\bibpunct[, ]{(}{)}{,}{a}{}{,}% Citation support using natbib.sty
\renewcommand\bibfont{\fontsize{10}{12}\selectfont}% Bibliography support using natbib.sty

\theoremstyle{plain}% Theorem-like structures

\theoremstyle{definition}

\theoremstyle{remark}

\begin{document}

\articletype{}

\title{Explainable GeoAI: Can saliency maps help interpret artificial intelligence's learning process? An empirical study on natural feature detection}

\author{
\name{Chia-Yu Hsu and Wenwen Li\thanks{CONTACT Wenwen Li. Email: wenwen@asu.edu}}
\affil{School of Geographical Sciences and Urban Planning, Arizona State University}
}

\maketitle

\begin{abstract}
Improving the interpretability of geospatial artificial intelligence (GeoAI) models has become critically important to open the “black box” of complex AI models, such as deep learning. This paper compares popular saliency map generation techniques and their strengths and weaknesses in interpreting GeoAI and deep learning models’ reasoning behaviors, particularly when applied to geospatial analysis and image processing tasks. We surveyed two broad classes of model explanation methods: perturbation-based and gradient-based methods. The former identifies important image areas, which help machines make predictions by modifying a localized area of the input image. The latter evaluates the contribution of every single pixel of the input image to the model’s prediction results through gradient backpropagation. In this study, three algorithms—the occlusion method, the integrated gradients method, and the class activation map method—are examined for a natural feature detection task using deep learning. The algorithms’ strengths and weaknesses are discussed, and the consistency between model-learned and human-understandable concepts for object recognition is also compared. The experiments used two GeoAI-ready datasets to demonstrate the generalizability of the research findings.
\end{abstract}

\begin{keywords}
XAI; artificial intelligence; deep learning; visualization; GeoAI
\end{keywords}

\section{Introduction}

Geospatial artificial intelligence (GeoAI) is an exciting and rapidly growing transdisciplinary research area that fuses AI with geographical laws and principles for solving geospatial problems in a data-driven manner \citep{li2020geoai}. Recent breakthroughs in deep learning technologies (e.g., convolutional neural networks [CNNs], recurrent neural networks, transformers, and deep reinforcement learning) have led to a flourishing of their applications in geographic information science (GIScience), such as terrain feature detection \citep{buscombe2018landscape,helber2019eurosat,li2020automated,wang2021geoai, hsu2021knowledge}, weather forecasting and nowcasting \citep{zhang2019application}, extreme climate event detection \citep{kurth2018exascale}, and neighborhood property quantification and change monitoring \citep{gebru2017using,koo2022how, li2022geoai2}. Although a wide variety of methods have been developed to support image analysis and machine vision, CNNs remain the most widely used type of deep learning model, achieving state-of-the-art performance \citep{li2022geoaia, li2022real}. By stacking multiple convolutional layers together, a CNN has the ability to hierarchically extract the prominent features of the target objects within an image scene for classification, detection, or segmentation purposes. 

Although CNNs have a well-defined model structure and reproducible parameters, their reasoning process remains a black box and is difficult to interpret due to the complex, non-linear nature of the model. The opaque decision-making process raises concerns about the scientific trustworthiness of the model's prediction results \citep{li2022geoai, kedron2021reproducibility}. The lack of model explainability may further hinder the replicability of research using GeoAI \citep{goodchild2021replication}. Hence, besides aiming to achieve outstanding performance in big data analytics in relation to aspects such as detection and prediction, it is also important to develop algorithms that increase our understanding of the knowledge-derivation process of an AI machine. To address this issue, the AI and GeoAI communities have been developing model explanation methods and tools to support the visual examination of model behavior and to connect this to the cognitive concepts used by humans when making decisions. For instance, feature attribution is a commonly used approach to explain a model’s predictions by attributing a decision to the input data. Based on different implementation mechanisms, the attribution can be identified at the pixel level, feature level, instance level, and concept level \citep{lundberg2017unified}. As the deep CNN models normally employ more complicated interaction strategies among the neural network layers, the most recent feature attribution methods have focused on identifying attribution at the first three levels \citep{simonyan2014deep,zeiler2014visualizing,fong2019understanding,selvaraju2020gradcam}. 

In the field of geography, efforts have also been made to enhance the interpretability and explainability of GeoAI. For instance, \citet{li2022extracting} employs an explainable AI (XAI) package, called SHapley Additive exPlanations (SHAP) to compare the spatial effects extracted by a machine learning-based regression model (i.e., XGBoost) with those of more traditional statistical approaches, such as the spatial lag model and multi-scale geographically weighted regression (MGWR) model. The research found that XGBoost has the ability to excerpt local spatial effects similar to those of classic models. This research shed new lights on using machine learning techniques for modeling spatial processes. \citet{xing2021integratinga} also used SHAP, within a land-use classification case, to understand which areas in the input data positively or negatively influenced a CNN model’s prediction result. Two example images were provided and the importance of feature maps at different stages of the convolution process were visualized. The authors argued that although the XAI tool can provide some level of explanation, it is difficult to link this explanation to a semantic or geographic concept. They therefore called for a deeper integration of XAI and GeoAI to understand how location and geographic attributes (e.g., those adhering to Tobler’s First Law of Geography) play a role in GeoAI modeling and decision making \citep{li2021tobler}. Another interesting work was published by \citet{duckham2022explainable}, who described an explainable spatiotemporal reasoning software framework for GeoAI applications. Different from the above works, this research adopted a top-down approach, using ontology and rule-based reasoning to derive new information. 

To further advance the field of explainable geospatial artificial intelligence (GeoAI), particularly as it relates to making deep learning models (e.g., CNNs) more explainable, this paper surveys and compares popular model explanation approaches from computer vision. It uses a GeoAI-ready natural feature dataset used in \cite{li2020automated} as an example to discuss the strengths and weaknesses of each method, catalog the feature types each method finds, and examine whether they agree with human-understandable concepts for characterizing a natural feature. To demonstrate our conclusions’ generalizability, the experiments were also conducted on GeoNat v1.0—another natural feature training dataset \citep{arundel2020geonat}. The remainder of this article is organized as follows: Section 2 presents a review of the literature. Section 3 introduces the feature attribution methods for XAI. Section 4 presents a series of experiments that identify the characteristics of each method and uses them to compare the model-learned features and human-understandable concepts for the image classification of natural features. Section 5 summarizes the findings and discusses the generalizability of the research findings. Section 6 concludes the paper and identifies future research directions.

\section{Literature Review}

Recently, deep learning techniques have shown outstanding predictive performance in image analysis and computer vision. However, they yield a lower level of explainability than other AI methods due to model complexity \citep{dw2019darpa,li2022geoai}. An initial attempt to observe and explain a deep learning model’s decision-making process was through visualizing various model components, such as filters and feature maps, but the information gained from this exercise can hardly be mapped to meaningful concepts \citep{li2017recognizing}. Recent XAI developments for deep learning models have shown two trends: developing global and per-decision explainable AI algorithms \citep{phillips2020four}. A global, explainable AI algorithm treats a deep learning model as a black box that can be queried and develops an algorithm approximation to explain the model. Testing with concept activation vectors \citep[TCAV;][]{kim2018interpretability} is one such global algorithm. It tests a model’s decision sensitivity to various pre-defined concepts, such as color, texture, and some specific patterns, from which it draws a conclusion about important decision factors for the model. Conversely, a per-decision explainable AI algorithm aims to determine why the model made a particular decision. Unlike the global algorithm, a per-decision algorithm does not require predefining or generating hypotheses about which concepts are important to the model. Instead, it distills evidence from each individual decision, allowing for the discovery of potentially unknown patterns and concepts to improve the understanding of a model’s learning process. We will mainly examine algorithms belonging to the second category in our research.

A very popular technique for deriving per-decision explanation is through the generation of a saliency map that visualizes the importance of different regions in the input image that factored into the final decision (i.e., classification or object detection). As known, deep CNN models stack multiple convolutional layers together to perform feature extraction from the original input image. During this process, downsampling is often applied to the input image and the feature maps generated from each convolutional layer. After the prominent image features are extracted, the feature maps are sent to fully connected layers for image classification. The model training process involves learning a set of model parameters, also called weights, to make a prediction as close to the ground truth as possible. The learning process is iterative by nature and contains forward propagation (i.e., forward pass) and backward propagation (i.e., backward pass). During the forward pass, the classification result is generated and compared with the ground-truth to calculate the loss. The loss is propagated back to each weight through backpropagation, which is a way to compute the partial derivative of the loss with respect to the weights such that a model learns how to adjust each weight and improve its predictability. The set of values capturing partial derivatives is called the \textit{gradient} and the weight adjustment process is called the \textit{gradient descent} because the weights are adjusted to \textit{minimize} the loss. 

The saliency map generation techniques leveraged to explain CNN model behaviors visually and quantitatively can be categorized into two broad classes: perturbation-based methods \citep{zeiler2014visualizing,ribeiro2016why,fong2017interpretable,sundararajan2017axiomatic,petsiuk2018rise,hesse2021fast} and gradient-based methods \citep{simonyan2014deep,zeiler2014visualizing,springenberg2015striving,zhou2016learning,selvaraju2020gradcam}. 

Perturbation-based methods systematically modify different portions of the input image and analyze the output sensitivity. They reveal which image regions contribute more and are likely more important to the model's prediction results. \citet{zeiler2014visualizing} developed an approach to perturb images with a gray patch sliding across the images and monitor result changes in both the classifier's output and the feature maps. We call this an \textit{occlusion approach}. Perturbation-based methods that involve strategies other than occluding images by raster scanning have also been introduced, such as one involving random masking using Monte Carlo sampling \citep{petsiuk2018rise} and another using superpixel masking \citep{ribeiro2016why}.  Here superpixels represent a continuous patch of pixels with similar intensities. \citet{fong2017interpretable} further proposed an optimization approach to identify a mask that minimizes the prediction score of a certain class; this mask would contain much of the information responsible for a classifier’s decision about an object class. In these perturbation-based techniques, the masks may consist of constant values, noise, or a blurring effect. Although perturbation-based methods have been proven to be effective in removing key information from the original image \citep{fong2017interpretable}, there is concern about whether these methods would also introduce spurious structures in the input image and therefore produce unexpected outputs \citep{nguyen2015deep,kurth2018exascale}.

Another family of methods for generating saliency maps is gradient-based methods. Unlike perturbation-based methods, which measure the influence of local areas on the prediction results, gradient-based methods measure the contribution of each individual pixel. These methods also answer the question of how much the prediction results would change if variation were introduced in a single given pixel. If pixels are viewed as variables and the output as a function of them, these methods are essentially computing the partial derivative of the output with respect to a given pixel. Backpropagation facilitates the calculation of partial derivatives, and the results can also be called the gradient, which can then be visualized as a \textit{saliency map}. It is for this reason that these types of CNN visualizations are called \textit{gradient-based methods}. Some recent methods calculate gradients at the input space \citep{simonyan2014deep,zeiler2014visualizing,springenberg2015striving}, while others need only gradients at the middle layers \citep{zhou2016learning,selvaraju2020gradcam}. For methods calculating gradients at the input space, the major difference is the way they handle backpropagation through the non-linear layers (e.g., Rectified Linear Unit [ReLU]). Different treatments will result in different saliency maps. \citet{simonyan2014deep} adopted vanilla backpropagation by which, at the non-linear layers, locations with negative values during the forward pass are recorded while gradients at those locations during the backward pass are suppressed. The deconvolution network \citep[DeconvNet;][]{zeiler2014visualizing} only propagates positive gradients and suppresses negative gradients. Finally, guided backpropagation \citep{springenberg2015striving} combines both the DeconvNet and the vanilla backpropagation, only propagating positive gradients at the locations with positive values during the forward pass. 

For the methods that need only gradients at a given middle convolutional layer, the gradients can be used as weights to combine feature maps at different channels and generate a saliency map. Because the feature maps are downsampled from the input image, this saliency map is considered as a downsampled version of the saliency map at the input space. To improve the visual effect, the saliency map is then upsampled to the same dimension as the input space. \citet{zhou2016learning} developed a Class Activation Map (CAM) by applying global average pooling \citep[GAP;][]{lin2014network} on the last convolutional layer and identifying the weights of feature maps from the following fully connected layer. \citet{selvaraju2020gradcam} proposed a generalization of a CAM called gradient-weighted CAM (Grad-CAM), which is applicable to different CNN models without using a GAP layer. The idea is to use gradient signals at the target layer as weights and prove them to be mathematically equal to the weights in CAM. In this way, Grad-CAM can derive the saliency map without needing to modify and retrain the CNN models. Methods that have followed up Grad-CAM include Grad-CAM++ \citep{chattopadhay2018gradcam} for the localization of multi-objects belonging to the same class and Score-CAM \citep{wang2020scorecam}, which obtains weights for corresponding feature maps without gradient calculation.

These gradient-based methods require only one forward-backward pass and are computationally efficient compared to the perturbation-based methods. However, the nonlinearity of the fully connected layers may cause gradient saturation or undesirable artifacts during backpropagation \citep{shrikumar2017learning}. Recent works have developed new strategies to address this issue. For example, \citet{sundararajan2017axiomatic} developed an integrated gradient method to aggregate the gradients over an entire image as it goes through continuous modifications, to avoid gradient saturation and unexpected artifacts. The entire image is gradually altered from the original image to an all-black image (i.e., baseline). The result is an aggregation of all the intermediate results after each alternation. This method can be considered as a combination of  perturbation-based and gradient-based methods and has shown promising results in many tasks \citep{sundararajan2017axiomatic}. However, as inferred from the algorithm strategy, the computation of integrated gradients requires repeated calculation over multiple iterations. To reduce computational cost, \citet{hesse2021fast} proposed a special class of CNNs to compute the result of integrated gradients with only one forward-backward pass. As a result, the saliency map itself can be used as an effective tool; for example, it can be integrated into regular training as priors \citep{erion2021improving,rieger2020interpretations}. 

Each of the discussed methods has its own strengths and weaknesses. For example, the perturbation-based method may underestimate the importance of features because other features already saturate the output \citep{shrikumar2017learning}. Gradient-based methods may generate the same results using an untrained model as those from a trained model, indicating that the method demonstrates only general model characteristics \citep{adebayo2018sanity}. Although CAM methods can generate a saliency map by passing through fewer non-linear layers and potentially suffer from fewer issues, they may predict inaccurate object locations due to the coarse resolution of the saliency map. In this paper, we explore the characteristics of different ``feature attribution-based" model explanation approaches and use two GeoAI-ready natural feature datasets in an image classification task to compare model-learned features with human-understandable features. In addition to interpreting a model’s reasoning process, we compare the strengths and weaknesses of these popular approaches and discuss ways to improve them within the growing field of explainable GeoAI.

\section{Methods}\label{sec_3}
Suppose we have a function $F: R^{w\times h} \rightarrow R^{m}$, which represents a CNN network for image classification. The input is an image $x\in R^{w\times h}$ with width $w$ and height $h$. The output is a vector of length $m$, representing the probability distribution $P$ over $m$ classes. $P^c$ stands for the classification probability of the class $c$ to the image $x$. For any given class $c$, we could generate an image $y$ where $y_{i,j}$ (i.e., pixel $[i,j]$ in image $y$) is the contribution of $x_{i,j}$ (i.e., pixel $[i,j]$ in image $x$) to the prediction of the image class $c$. We call this image $y$ a saliency map. In this section, we describe in greater detail the three different methodologies adopted to generate a saliency map. 

\subsection{Perturbation-based saliency map generation (i.e., occlusion method)}
To generate a perturbation-based saliency map, we adopt the occlusion method from the work by \citet{zeiler2014visualizing}. Given the function $F$ and an input image $x$, we calculate the output probability distribution $P$. $P^c$ represents the output probability for the class $c$.  Next, we create a black patch at a given size. The patch slides through the image $x$ in a row-prime scan order with a stride step of 1. At each step, a new image $x\prime(i,j)$ is generated by replacing a portion of the image $x$ with the black patch. The center of the patch is located at the pixel $(i,j)$. The new image $x\prime(i,j)$ is then fed into the model $F$, and the new probability distribution $P\prime(i,j)$ is calculated. $P\prime^c(i,j)$ represents the new probability of the class $c$. The difference between the two probabilities, $P^c$ and $P\prime^c(i,j)$, shows the importance of the information in the black patch to the model's prediction of the class c: 
\begin{equation}\label{eqn_1}
    \text{Occlusion}^c_{i,j}=P^c-P\prime^c(i,j)
\end{equation}
Finally, by calculating $\text{Occlusion}^c_{i,j}$ over all pixels, we can generate the saliency map indicating which pixels or pixel regions are more or less important for predicting the existence of the class $c$ in image $x$.

\subsection{CAM-series saliency map generation with Grad-CAM}
The idea of Grad-CAM \citep{selvaraju2020gradcam} is to calculate the gradient at the last convolutional layer in the network from a given class-prediction score, together with the feature maps at this layer, to generate a coarse saliency map (i.e., a CAM) highlighting important areas for predicting the class. Grad-CAM can be directly applied to a trained CNN network to derive a saliency map mathematically. It allows for the generalization of a regular CAM \citep{zhou2016learning} without the need to add a GAP (Global Average Pooling) layer to the network and retrain the CNN model. The process of generating a saliency map with a regular CAM is as follows. Given the function $F$ and an input image $x$, $A^k_{i,j}$ can be used to represent the value at location $(i,j)$ of the $k$th feature map resulting from the last convolutional layer. Further, $g^k$ is the output of GAP, the function of which can be written as:
\begin{equation}\label{eqn_2}
    g^k=\frac{1}{Z}\sum_{i,j}A^k_{i,j}
\end{equation}
where $Z$ is the total number of pixels in feature map $A^k$.

Next, let $S^c$ represent the prediction score of the class $c$. $S^c$ can be derived by calculating the composited value from $g^k$, as follows:
\begin{equation}\label{eqn_3}
    S^c=\sum_kw^{ck}g^k
\end{equation}
where $w^{ck}$ is the relative weight of $g^k$ to $S^c$. Accordingly, $w^{ck}$ can indicate the importance of $g^k$ to the prediction of the class $c$ because the bigger $w^{ck}$ is, the more weight that $g^k$ requires to derive the class $c$. The probability of an image being the class $c$, $P^c$, is derived by applying softmax to score $S$ over all classes. Softmax is a normalization function to transfer the prediction scores to a probability distribution over the predicted classes. The higher $S^c$ (i.e., class prediction score) is, the higher $P^c$ will be. Next, by substituting $g^k$ in equation \ref{eqn_3} with its definition in equation \ref{eqn_2}, we can derive the following:
\begin{equation}\label{eqn_4}
    S^c=\sum_kw^{ck}\frac{1}{Z}\sum_{i,j}A^k_{i,j}=\frac{1}{Z}\sum_{i,j}\sum_kw^{ck}A^k_{i,j}
\end{equation}
Further, by making
\begin{equation}\label{eqn_5}
    M^c_{i,j}=\sum_kw^{ck}A^k_{i,j}
\end{equation}
we have
\begin{equation}\label{eqn_6}
    S^c=\frac{1}{Z}\sum_{i,j}M^c_{i,j}
\end{equation}
Because $S^c$ is the prediction score for the class $c$, $M^c_{i,j}$ directly indicates the importance of the location $(i,j)$ in the last feature map (i.e., by aggregating information from all channels) to the class $c$. Therefore, if we calculate $M^c_{i,j}$ over all locations, we can derive a saliency map indicating the important areas for making predictions about the class $c$. 

However, to generate $M^c_{i,j}$ (in equation \ref{eqn_5}) over all locations, we need both $A^k_{i,j}$ and weight $w^{ck}$ at the pixel level. In a regular CAM approach \citep{zhou2016learning}, $w^{ck}$ is generated through GAP, which requires changes to the CNN model architecture. In Grad-CAM \citep{selvaraju2020gradcam}, however, a generalization of the CAM is achieved through the mathematical transformation of the gradient calculations. The gradients are the partial derivatives of the target class score $S^c$ with respect to the feature map $k$). Mathematically, it can be written as follows. First, by taking partial derivatives of $S^c$ with respect to $g^k$ from equation \ref{eqn_3} and combining information from equation \ref{eqn_2}, we can derive
\begin{equation}\label{eqn_7}
    \frac{\partial S^c}{\partial g^k}=w^{ck}=\frac{\partial S^c}{\partial A^k_{i,j}}\cdot\frac{\partial A^k_{i,j}}{\partial g^k}=\frac{\partial S^c}{\partial A^k_{i,j}}\cdot Z
\end{equation}
For the weight $w^{ck}$, its relation to a given pixel at $(i,j)$ in feature map $A^k$ can be identified using
\begin{equation}\label{eqn_8}
    w^{ck}=Z\cdot\frac{\partial S^c}{\partial A^k_{i,j}}
\end{equation}
If we calculate the summation of $w^{ck}$ over all pixels, it can be expressed as
\begin{equation}\label{eqn_9}
    \sum_{i,j}w^{ck}=\sum_{i,j}Z\cdot\frac{\partial S^c}{\partial A^k_{i,j}}=Z\cdot\sum_{i,j}\frac{\partial S^c}{\partial A^k_{i,j}}
\end{equation}
At the same time, the weight $w^{ck}$ is not a function of the location $(i,j)$, so the summation can also be expressed as 
\begin{equation}\label{eqn_10}
    \sum_{i,j}w^{ck}=Z\cdot w^{ck}
\end{equation}
Combining equation \ref{eqn_9} and equation \ref{eqn_10}, the weight $w^{ck}$ can be mathematically derived from the partial derivatives of $S^c$ with respect to all pixels at feature map $A^k$, as follows:
\begin{equation}\label{eqn_11}
    w^{ck}=\sum_{i,j}\frac{\partial S^c}{\partial A^k_{i,j}}
\end{equation}
Equation \ref{eqn_11} indicates a way to calculate $w^{ck}$ without using a GAP layer. Therefore, we can replace $w^{ck}$ in equation \ref{eqn_5} with equation \ref{eqn_11}, giving
\begin{equation}\label{eqn_12}
    \text{GradCAM}^c_{i,j}=M^c_{i,j}=\sum_k\sum_{i,j}\frac{\partial S^c}{\partial A^k_{i,j}}\cdot A^k_{i,j}
\end{equation}
which provides the importance of a pixel at a given location $(i,j)$ to the prediction of the class $c$ based on the class prediction score $S^c$ and the final feature map $A$ at each channel $k$. This way, no architecture changes to the model are required. Because the generated saliency map is a coarse map with a resolution smaller than the original input image, it can also be upsampled to obtain a better visual effect. 

\subsection{Gradient-based saliency map generation with integrated gradients}
The motivation of integrated gradients \citep{sundararajan2017axiomatic} is to solve the challenge of separating actual errors from misbehavior of the model and errors caused by the model explanation method. The authors identified two axioms in order to evaluate the validity of different attribution methods: sensitivity and implementation invariance. The sensitivity axiom indicates that a pixel should be given a non-zero contribution if two input images have different predictions but only differ in that pixel. The implementation invariance axiom indicates that the saliency maps are always identical for two functionally equivalent models. \textit{Functionally equivalent} models are those yielding the same output given any input, despite their different implementations. By satisfying the two axioms, errors introduced by the saliency map generation method can be disregarded. Using the two axioms, a new method -- the integrated gradients method -- was developed \citep{sundararajan2017axiomatic}. One key concept of the integrated gradients method is the use of a \textit{baseline} image. The idea of a \textit{baseline} image is implicitly used in previously derived methods because, when assigning credit to a pixel or a sub-region of an image, we actually consider the image without the pixel or the sub-region as a baseline image with which to compare the difference in outputs. In the integrated gradients method, however, the baseline image is explicitly used, taking the form of a black image (i.e., all zeros) for a classification task. Given the classification function $F$, an input image $x$, and a baseline image $x\prime$, the integrated gradients method defines the contribution of a pixel $x_{i,j}$ to the prediction of a class $c$ as
\begin{equation}\label{eqn_13}
    \text{IntegratedGradients}^c_{i,j}=(x_{i,j}-x\prime_{i,j})\times\int_{\alpha=0}^1\frac{\partial F(x\prime+\alpha\cdot(x-x\prime))^c}{\partial x_{i,j}}d\alpha
\end{equation}
Equation \ref{eqn_13} demonstrates why this method is named after \textit{integrated gradients}, as it computes and integrates gradients of the final output with respect to pixel $x_{i,j}$ from the baseline image $x\prime$ to the original input image $x$. In other words, it identifies the “path integrals” of gradients along the linear path from $x\prime$ to $x$. Within the fundamental theorem of path integrals, if the function $F$ is differentiable almost everywhere, the summation of equation \ref{eqn_13} over all pixels can be written as 
\begin{equation}\label{eqn_14}
    \sum_{i,j}\text{IntegratedGradients}^c_{i,j}=F(x)^c-F(x\prime)^c
\end{equation}
In our experiment, we assume the black baseline image would make the prediction score near zero -- that is, $F(x\prime)^c\approx0$. Therefore, the prediction of image $x$ being the class $c$ ($F(x)^c$) equals the summation of the contributions from all pixels. This confirms the accuracy of equation \ref{eqn_13} in calculating an individual pixel’s contribution to the final prediction. The integrated gradients method also satisfies the aforementioned axioms. In terms of sensitivity, assuming that only one given pixel is different between the input image $x$ and the baseline $x\prime$, from equation \ref{eqn_13}, we know that all of the integrated gradients would be zero except for those of the given pixel. From equation \ref{eqn_14}, the contribution of this given pixel becomes the difference between the outputs of $F(x)^c$ and $F(x\prime)^c$. Therefore, if $F(x)^c$ and $F(x\prime)^c$ are different, the contribution is non-zero. In terms of implementation invariance, because equation \ref{eqn_13} only considers the gradients of the model, it is invariant to different architectures. To compute integrated gradients more efficiently, instead of calculating the integrals from equation \ref{eqn_13}, we can also add the gradients at sufficiently small intervals from the baseline image $x\prime$ to the original image $x$. An approximation via summation can be achieved as 
\begin{equation}\label{eqn_15}
    \text{IntegratedGradients}^c_{i,j}\prime=(x_{i,j}-x\prime_{i,j})\times \sum^m_{k=1}\frac{\partial F(x\prime+\frac{k}{m}\cdot(x-x\prime))^c}{\partial x_{i,j}}\cdot\frac{1}{m}
\end{equation}
where $m$ is the number of intervals. The authors \citep{sundararajan2017axiomatic} suggested that a number between 20 and 300 steps is sufficient to approximate the integral (at a 95\% accuracy). 

\section{Experimental Results}
\begin{figure}
    \makebox[\textwidth][c]{
        \subcaptionbox{Crater}[.3\textwidth]{\includegraphics[width=.3\textwidth]{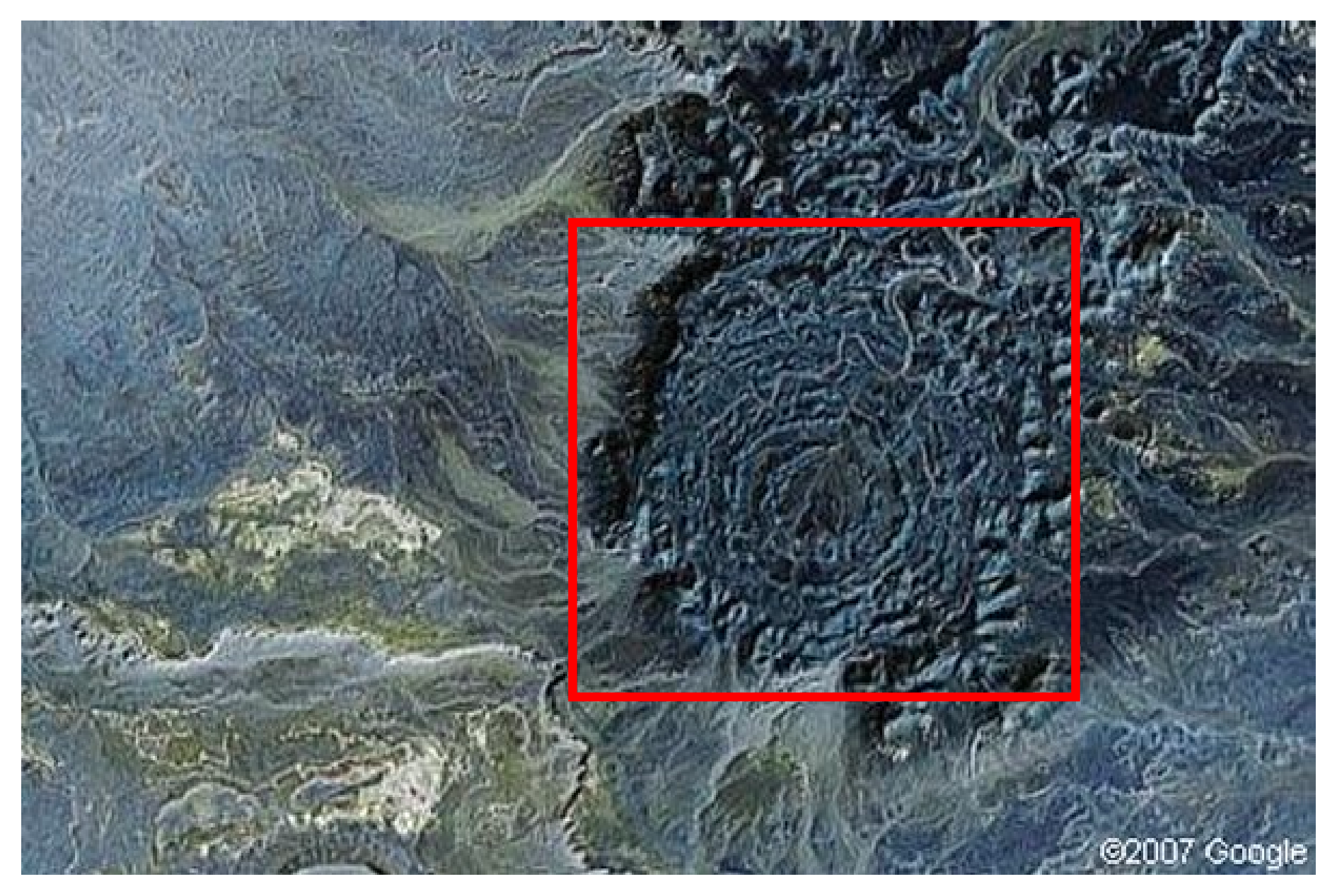}}
        \subcaptionbox{River}[.3\textwidth]{\includegraphics[width=.3\textwidth]{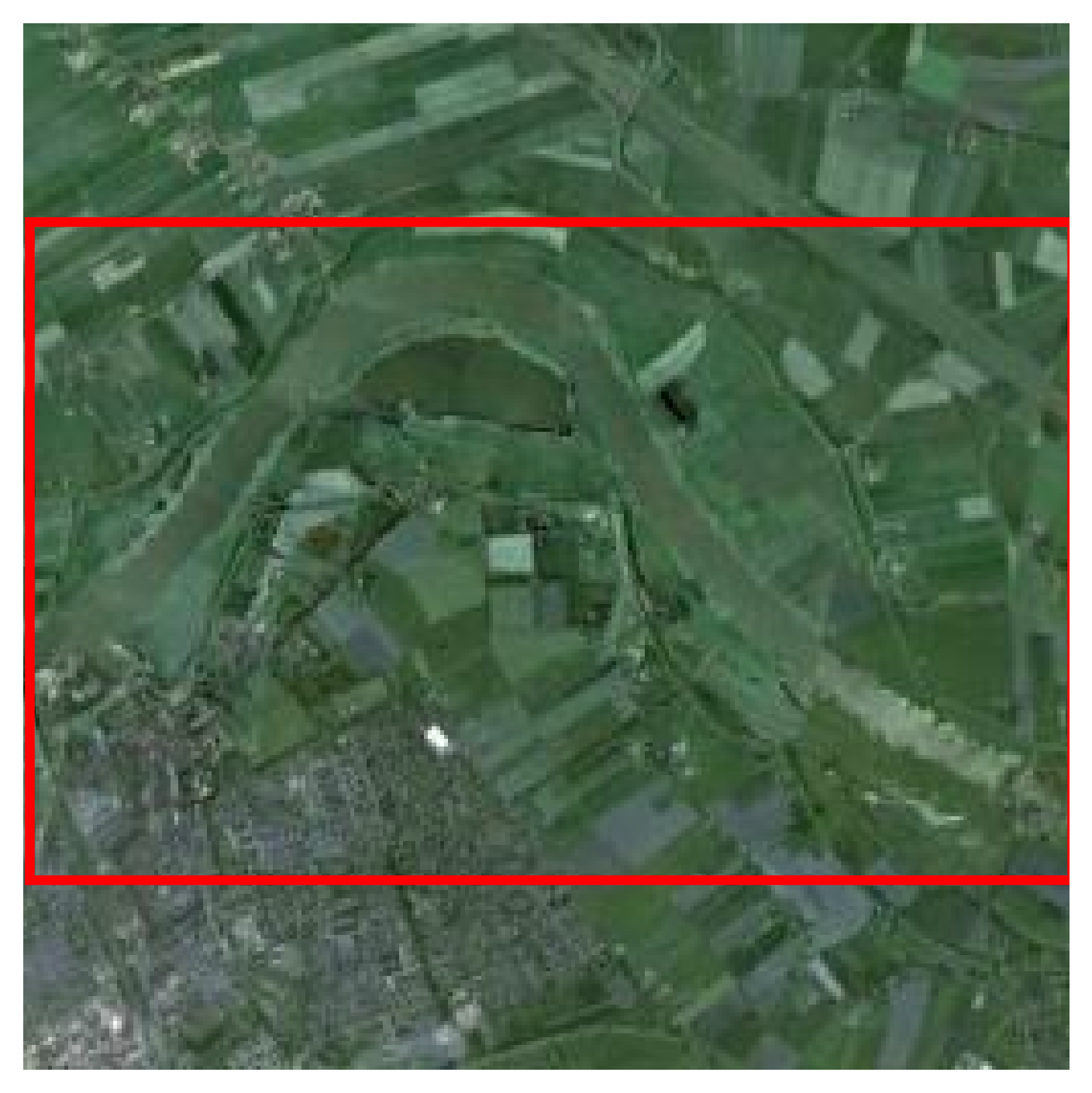}}
        \subcaptionbox{Meander}[.3\textwidth]{\includegraphics[width=.3\textwidth]{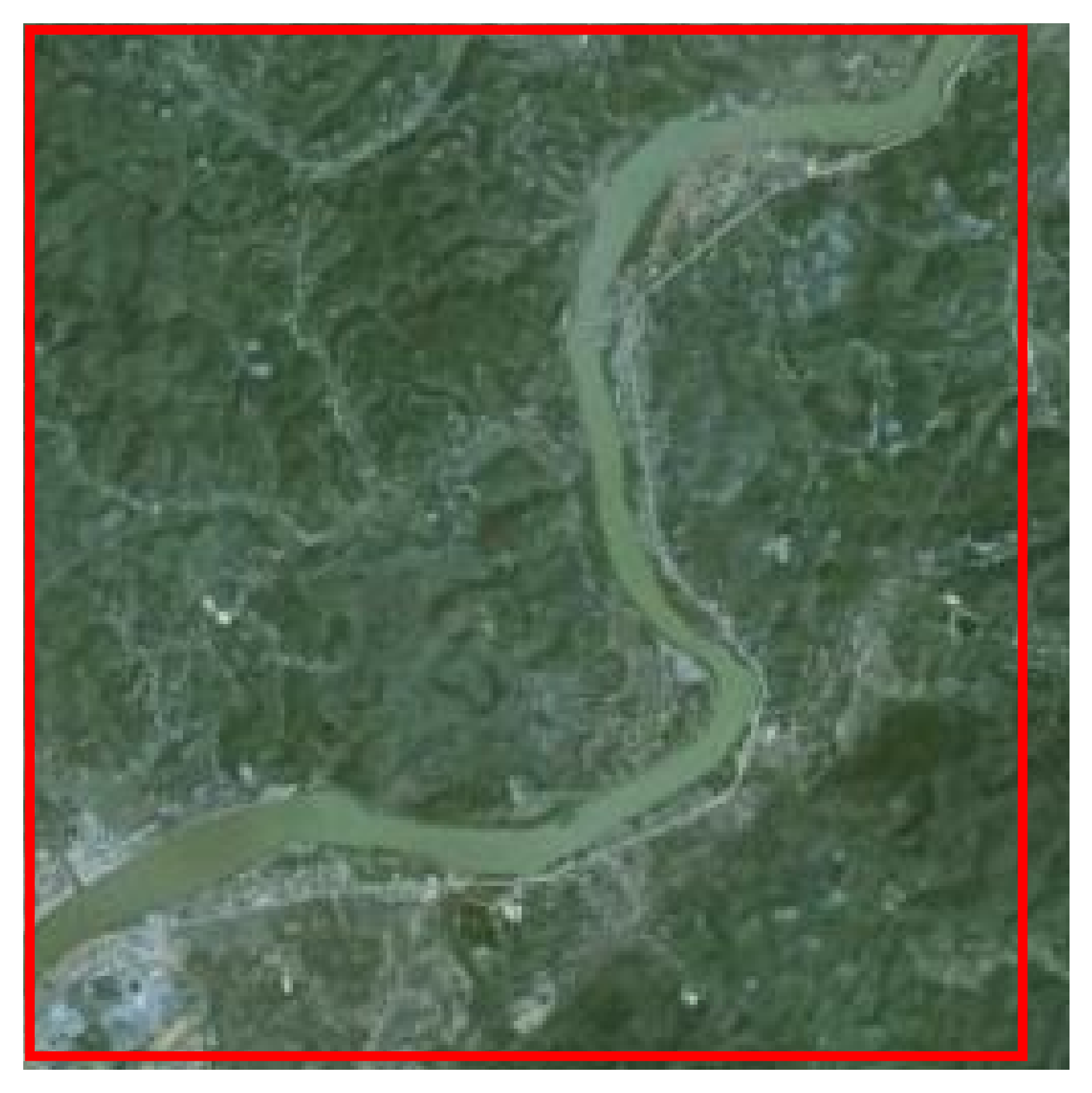}}
        \subcaptionbox{Iceberg tongue}[.3\textwidth]{\includegraphics[width=.3\textwidth]{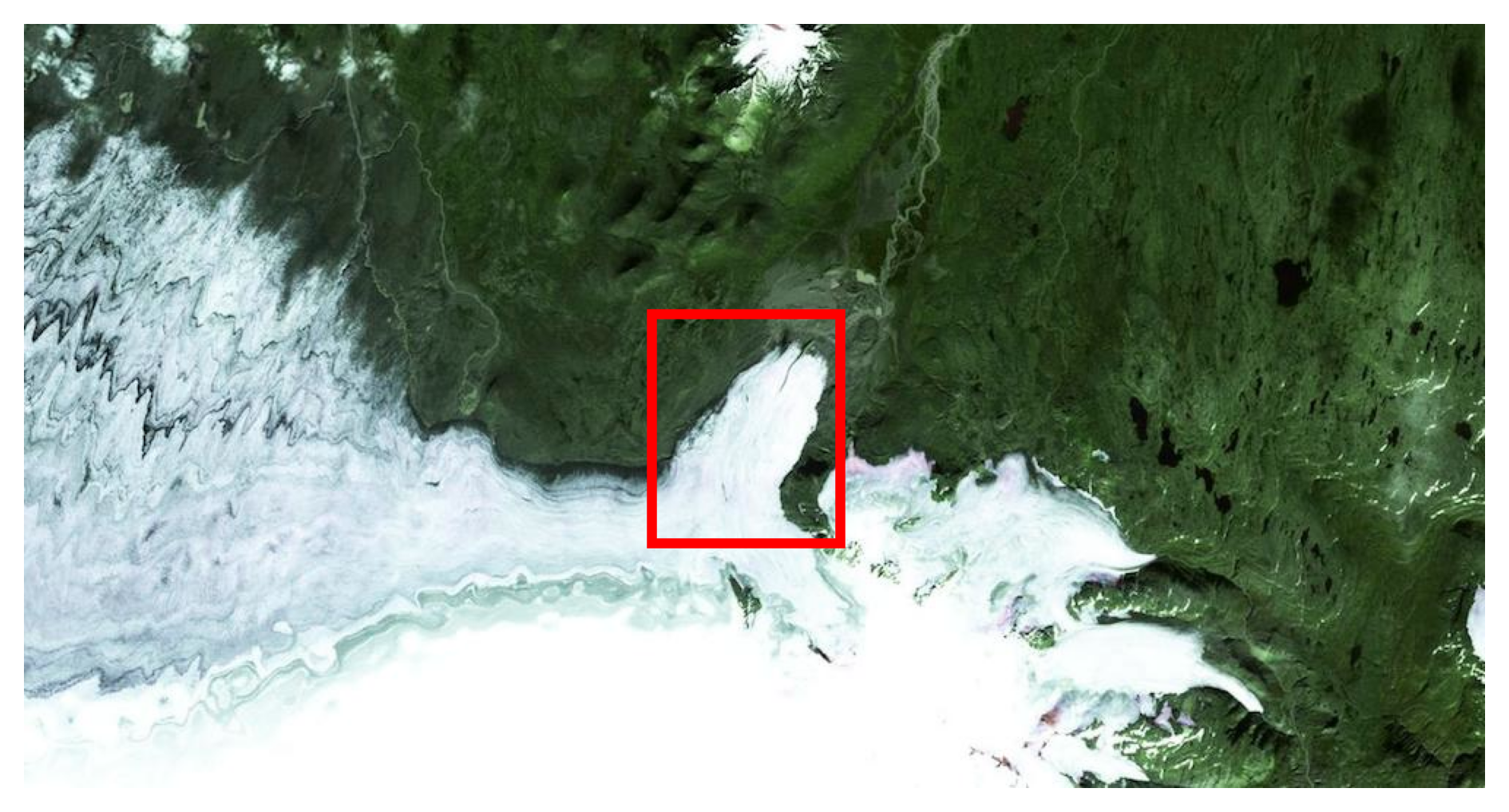}}
    }
    \makebox[\textwidth][c]{
        \subcaptionbox{Lake}[.3\textwidth]{\includegraphics[width=.3\textwidth]{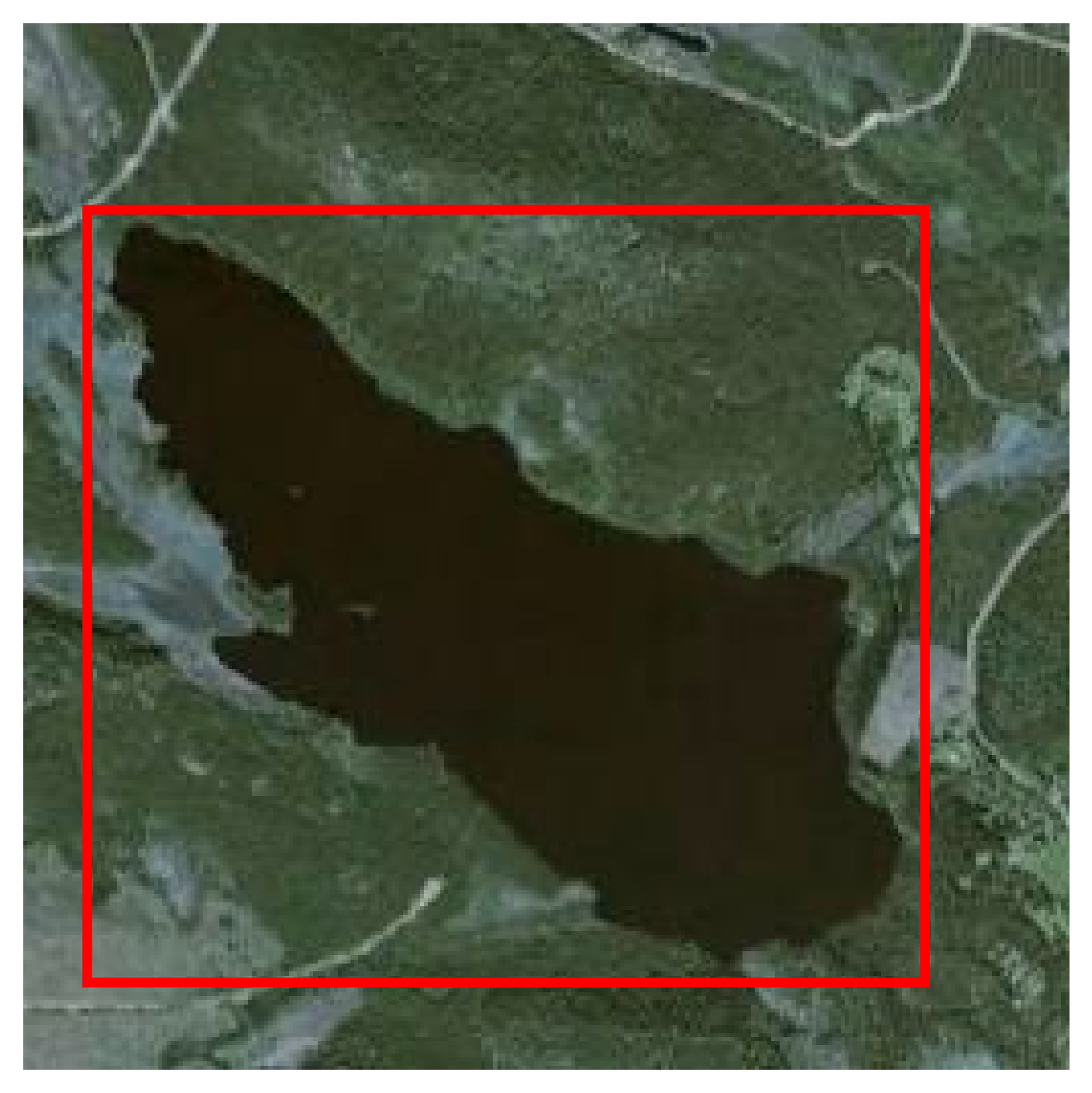}}
        \subcaptionbox{Volcano}[.3\textwidth]{\includegraphics[width=.3\textwidth]{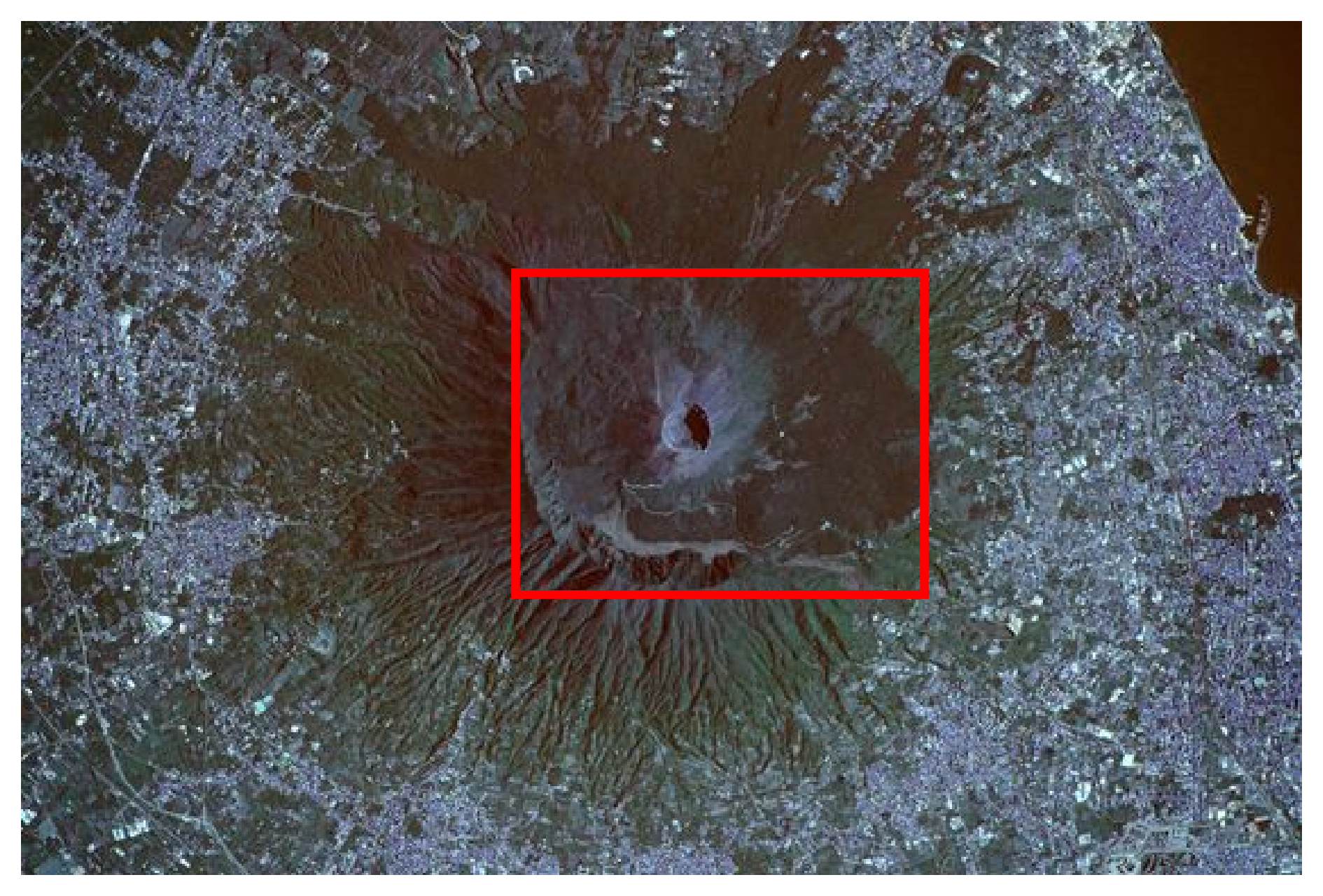}}
        \subcaptionbox{Hill}[.3\textwidth]{\includegraphics[width=.3\textwidth]{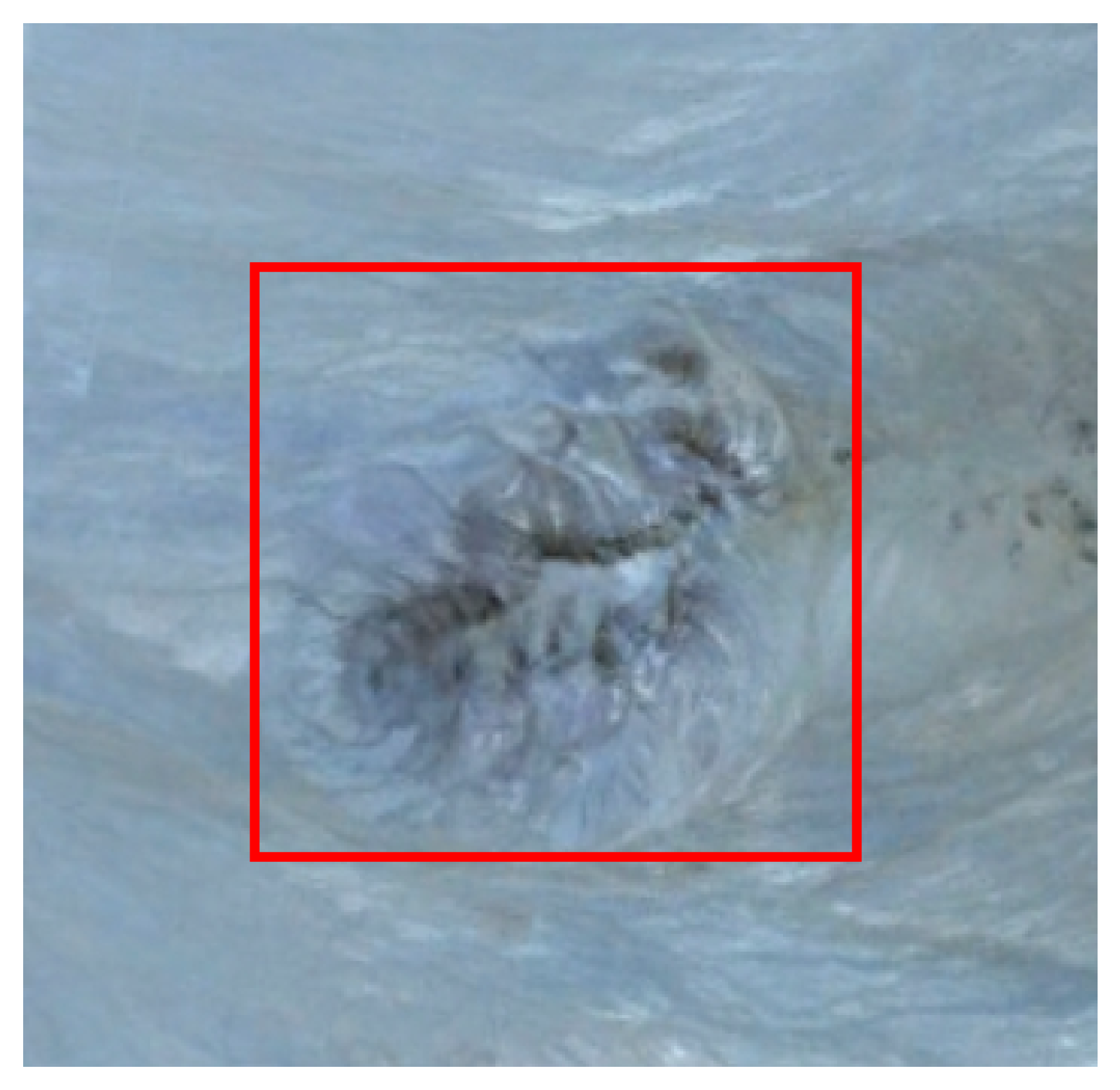}}
        \subcaptionbox{Sand dunes}[.3\textwidth]{\includegraphics[width=.3\textwidth]{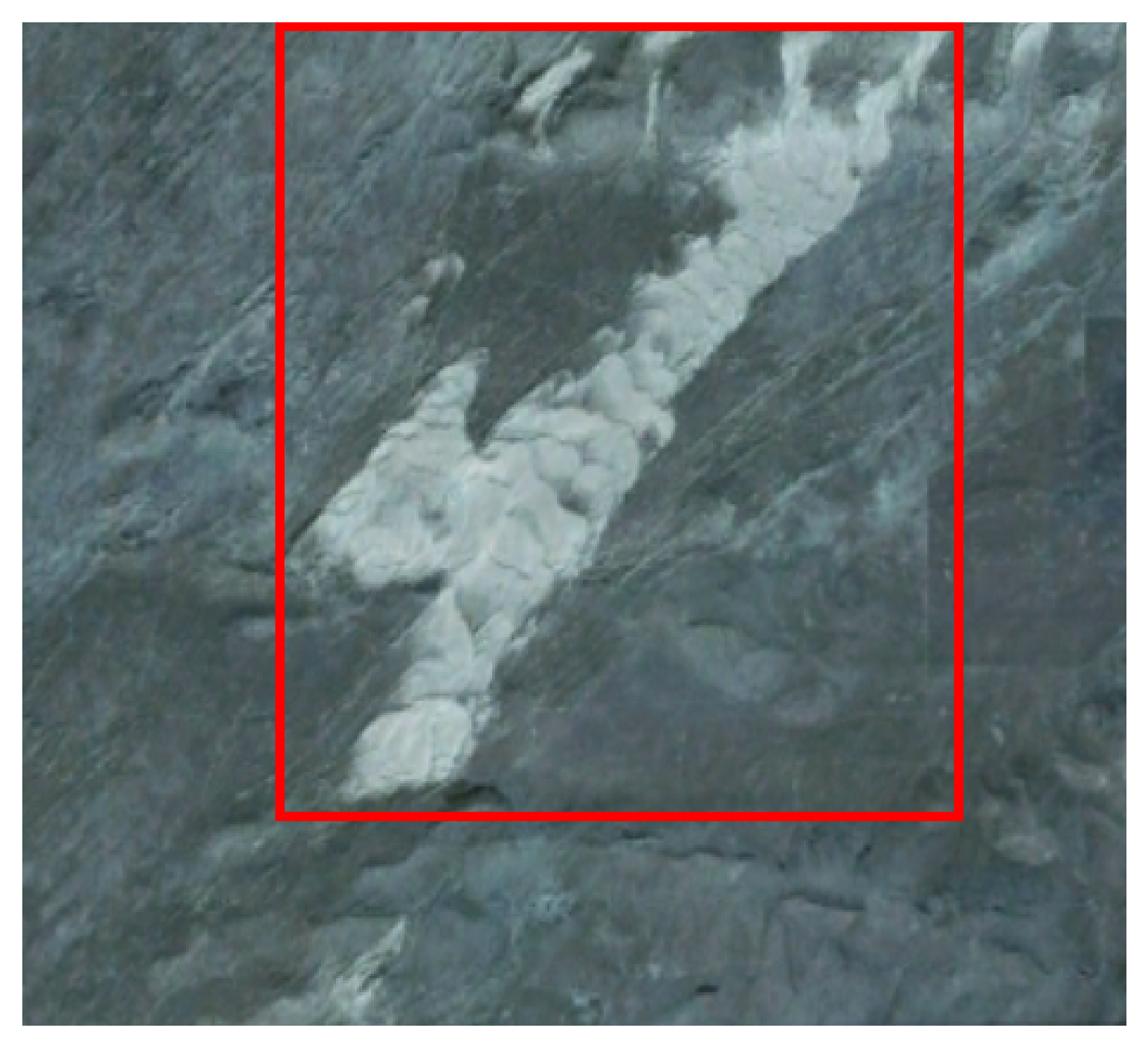}}
    }
    \caption{Sample dataset images. Red boxes are labels indicating object locations.}
    \label{fig_1}
\end{figure}

To understand how the CNN models make decisions in image analysis tasks specifically when inspecting the different types of terrain features in an image, we adopted the same dataset used in the work by \citet{li2020automated}. The terrain dataset contained 826 natural features of eight types (i.e., crater, meander, river, hill, lake, volcano, iceberg tongue, and sand dune). Each category had 100 to 108 images. Figure \ref{fig_1} shows examples of training images containing features of each type. Briefly, a crater is “a circular-shaped depression on the surface of the land” \citep{mcewen1983usgs}. A river is “a natural flowing watercourse, usually freshwater, flowing towards an ocean, sea, or another river” \citep{wiki:River}. A meander is “a winding curve or bend in a river” \citep{wiki:Meander}. An iceberg tongue is “a seaward extension of a glacier tongue, which consists of floating glacier ice that is still connected [to the glacier] and extends to the sea or ocean from a glacier” \citep{herzfeld2004atlas}. A lake is “an area of land that is filled with water” \citep{purcell_2018}. A volcano is “a rupture in the crust of a planetary-mass object, such as Earth, that allows hot lava, volcanic ash, and gases to escape from a magma chamber below the surface” \citep{wiki:Volcano}. A hill is “a naturally raised area of land, not as high or craggy as a mountain” \citep{hill_ngs}. A sand dune is “a small ridge of hill of sand found in a desert or on top of a beach” \citep{EarthEclipse_2022}. 

The dataset was originally used for object detection tasks, and here, we adopted it for image-level classification using only the class labels (i.e., without the bounding-box labels) for model training and prediction. The dataset was randomly separated into 70\% for training and 30\% for testing. The classification model was trained using a pretrained VGG16 CNN \citep{simonyan2014very}. The final training accuracy was 99.83\%, and the testing accuracy was 88.05\%. All experiments were conducted on the Amazon EC2 platform. The g4dn.xlarge instance, with a NVIDIA T4 graphics processing unit (GPU) that had a 16GB memory, was used to run the experiments. 

In each experiment, we applied three model explanation techniques -- including occlusion sensitivity by \citet{zeiler2014visualizing}, Grad-CAM by \citet{selvaraju2020gradcam}, and the integrated gradients method by \citet{sundararajan2017axiomatic} -- to explain the model’s decision-making process. The three methods were all applied to a trained model, and the generation of the saliency maps is detailed in Section \ref{sec_3}. By comparing the results, we aimed to examine the characteristics of each method, cross-validate their findings, and, most importantly, use them jointly to explain the model’s classification results in image analysis and natural feature recognition. Figures \ref{fig_2} - \ref{fig_7}, below, provide the results. In these figures, the first column displays the original image, with the targeted feature(s) highlighted in a rectangle. The ground-truth label of the object class and the model's prediction results, with a confidence score, are listed below the image. The second column lists the saliency maps generated from the three model explanation methods. The last column shows the blending result of the saliency map on top of the original image to make observing highlighted regions easier. For each saliency map,  higher values (indicated in red) represent the pixels or image areas that are more important to the prediction results. After cross-validating the saliency maps, we found that the findings of each method showed distinct characteristics due to the methods' unique model interpretation strategies. 

\subsection{Ability to identify multiple objects of the same type}
\begin{figure}
    \captionsetup[subfigure]{justification=centering, skip=-2pt}
    \centering
    \begin{minipage}{0.3\textwidth}
        \begin{subfigure}[]{\textwidth}
            \includegraphics[width=\textwidth]{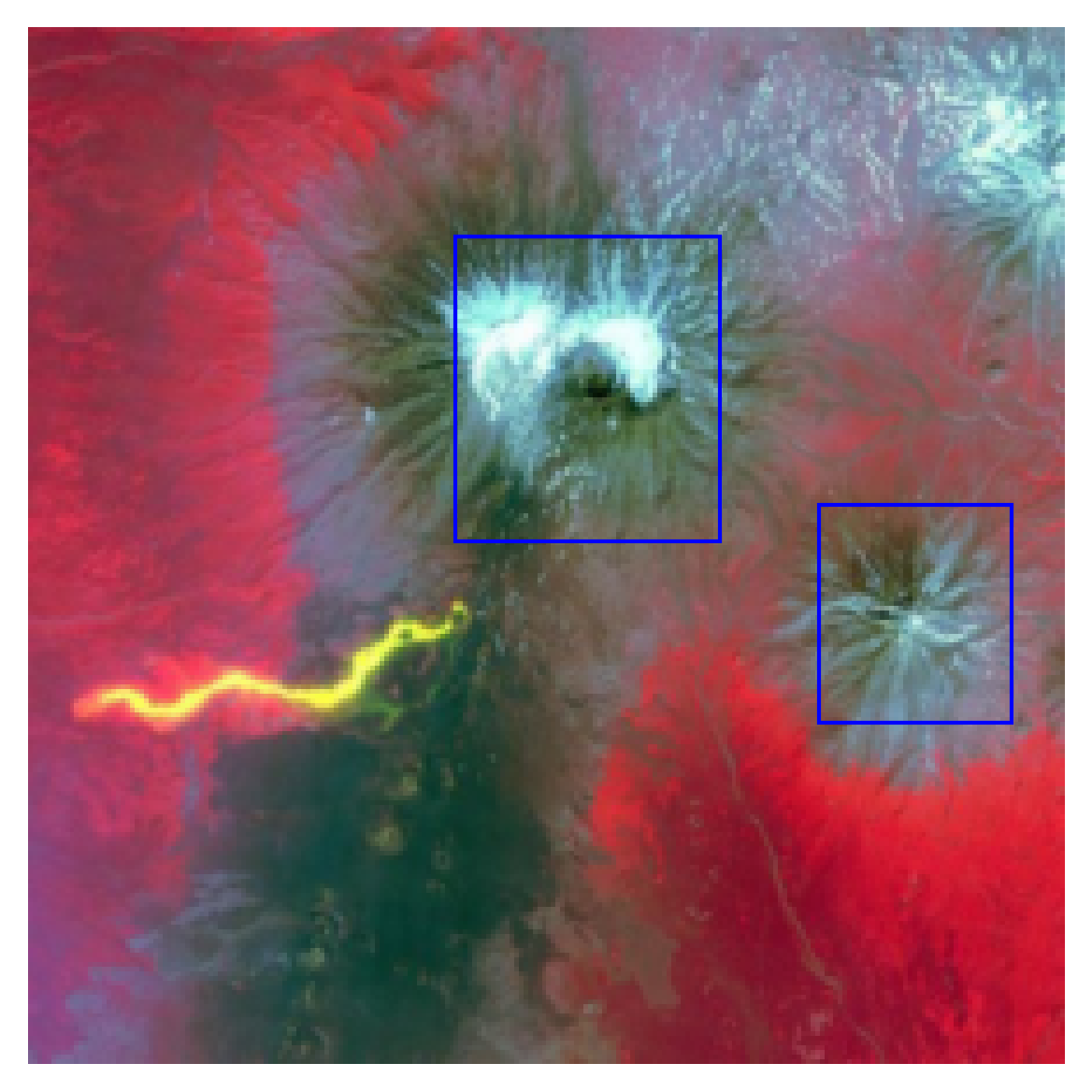}
            \caption*{Label: volcano (1.00)\\Prediction: volcano (1.00)}
        \end{subfigure}
    \end{minipage}
    \begin{minipage}{0.65\textwidth}
        \begin{subfigure}[]{\textwidth}
            \includegraphics[width=\textwidth]{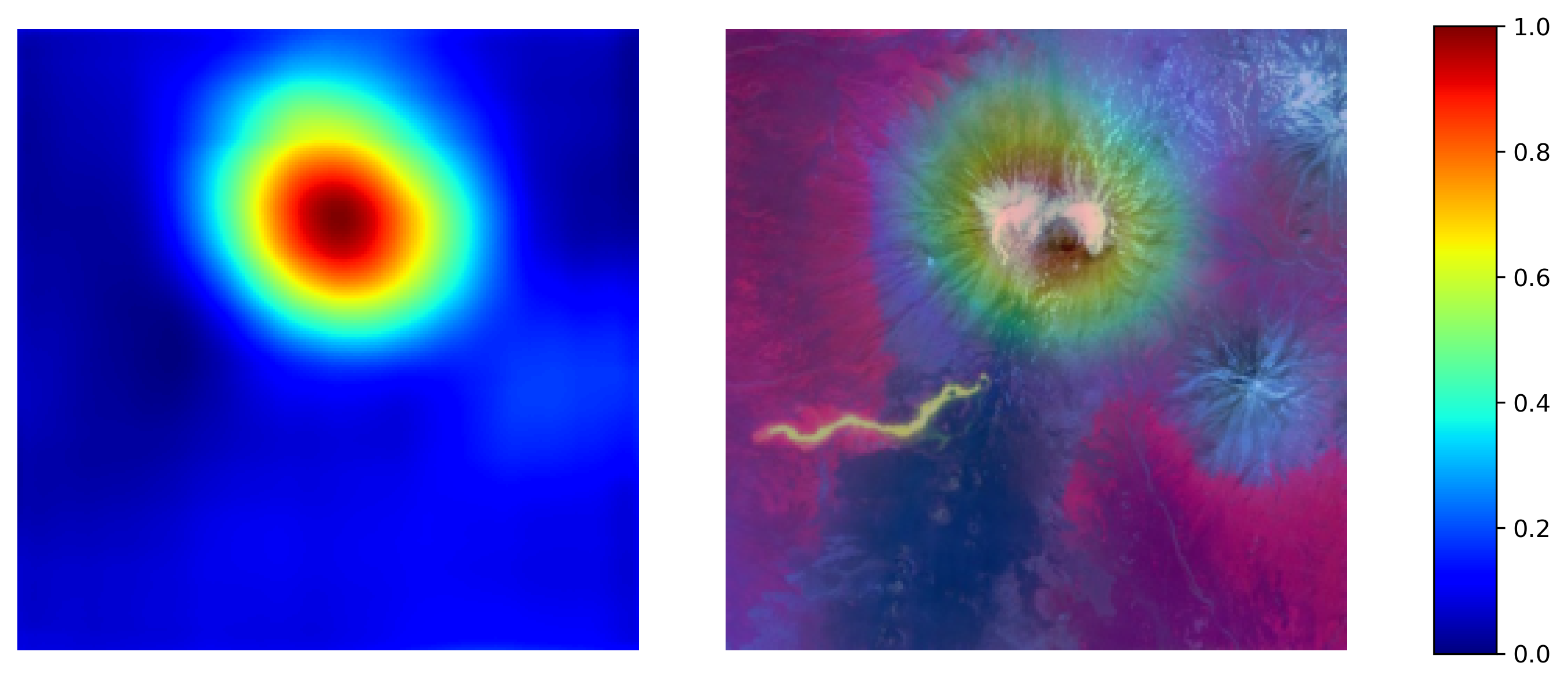}
            \caption{Occlusion}
        \end{subfigure}
        \begin{subfigure}[]{\textwidth}
            \includegraphics[width=\textwidth]{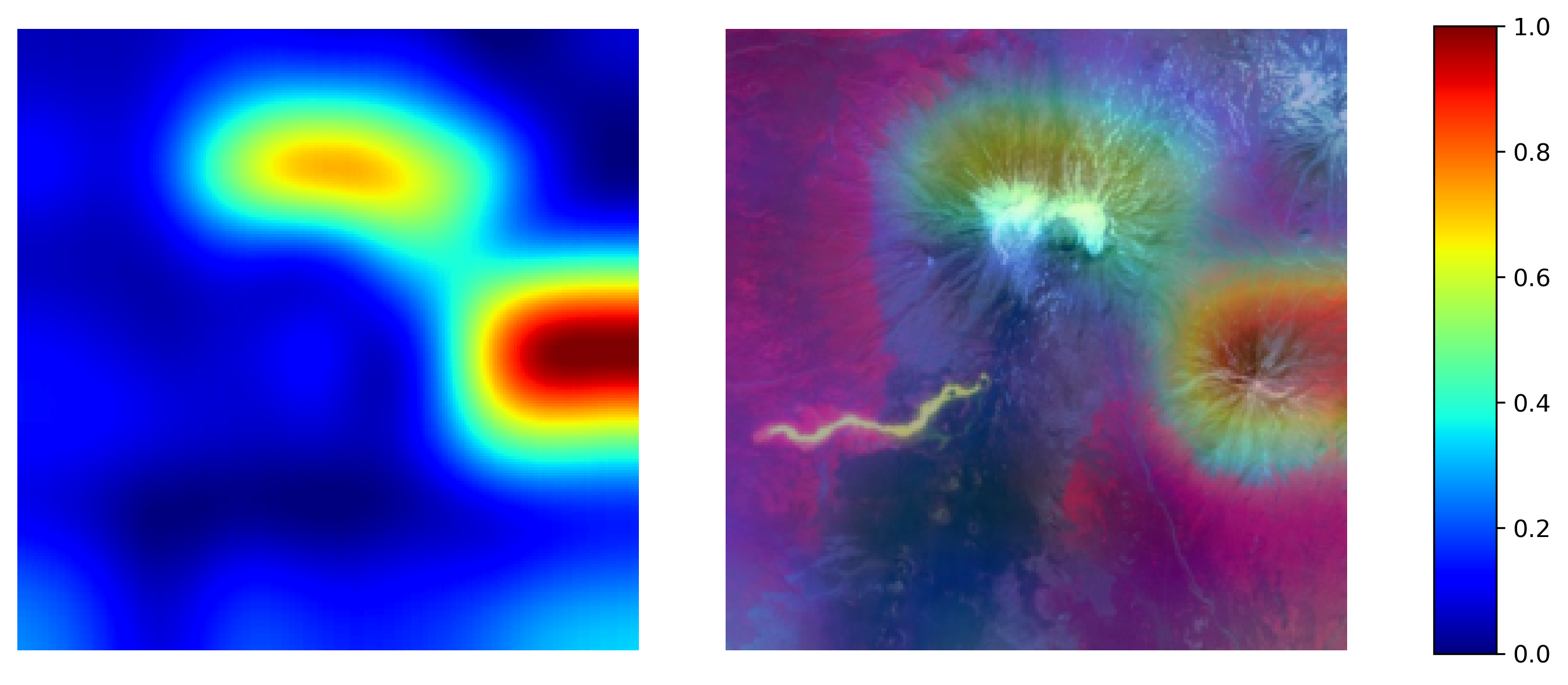}
            \caption{Grad-CAM}
        \end{subfigure}
        \begin{subfigure}[]{\textwidth}
            \includegraphics[width=\textwidth]{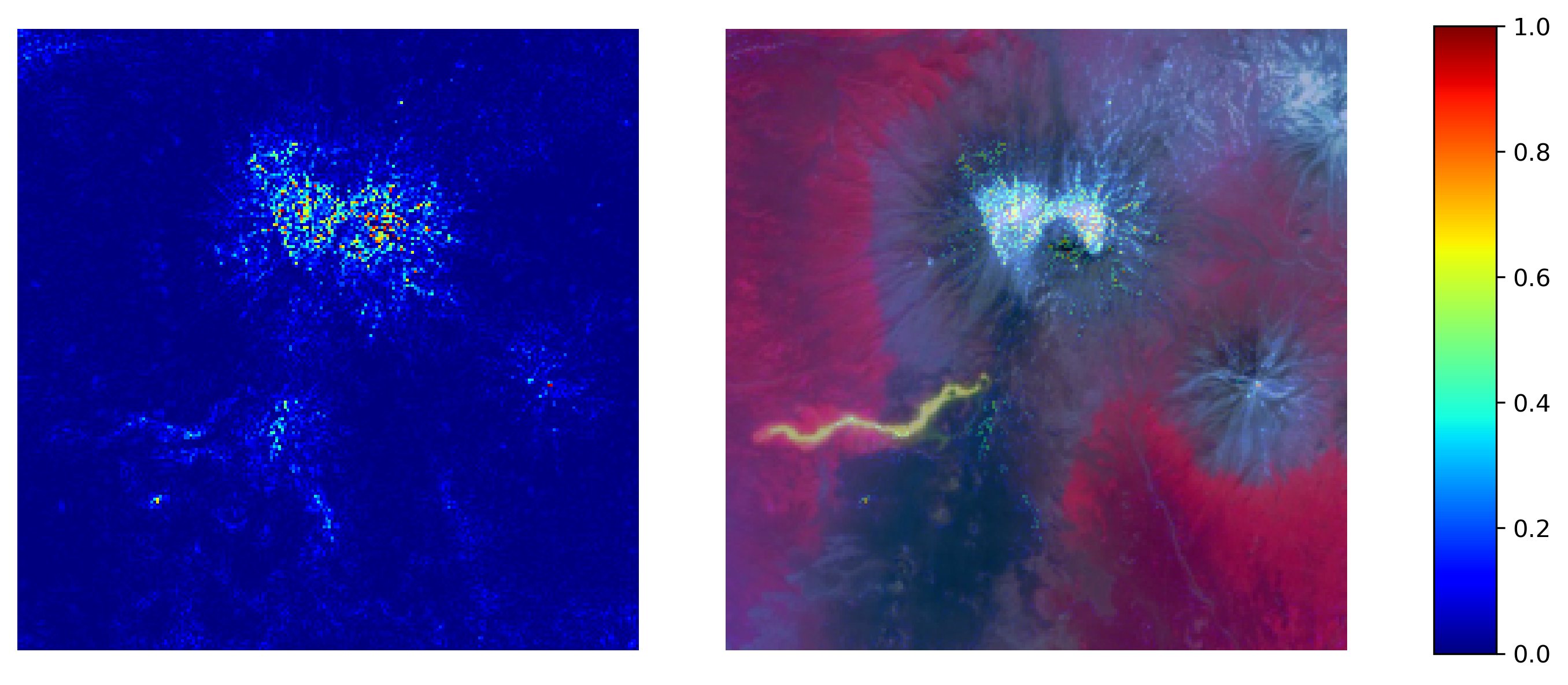}
            \caption{Integrated Gradients}
        \end{subfigure}
    \end{minipage}
    \caption{Saliency maps showing important image areas identified by the deep learning model. First column: original image with ground-truth and model prediction results. Second column: saliency map generated from (a) the occlusion method, (b) Grad-CAM, and (c) integrated gradients (Red: high value, Blue: low value). Third column: saliency map overlaid on the original image.}
    \label{fig_2}
\end{figure}

Figure \ref{fig_2} presents an image with multiple occurrences of objects belonging to the same class (i.e., volcanoes). According to the generated saliency maps, both Grad-CAM and integrated gradients were able to highlight multiple objects belonging to the same class, while the occlusion method highlighted only a single object. This is due to the limitations of the occlusion method in generating accurate visual results. Occlusion continuously replaces part of the input image with a black patch and calculates the change of the output probability. However, if there are multiple objects belonging to the same class in the image and one of them already saturates the output, the blocking of other recessive objects will not change the output probability. This result shows that the occlusion method cannot fully reflect how the model processes images containing multiple objects of the same class. Compared to the occlusion method, Grad-CAM combines multiple feature maps, wherein each map could contain features from different objects. The integrated gradients method adopts pixel-wise evaluation and integrates all results from the baseline image to the input image to avoid output saturation by a certain object. The results also illustrate the importance and value of using joint evaluation instead of a single approach.  

\subsection{Ability to identify multiple prominent features of the same object}
\begin{figure}
    \captionsetup[subfigure]{justification=centering, skip=-2pt}
    \centering
    \begin{minipage}{0.3\textwidth}
        \begin{subfigure}[]{\textwidth}
            \includegraphics[width=\textwidth]{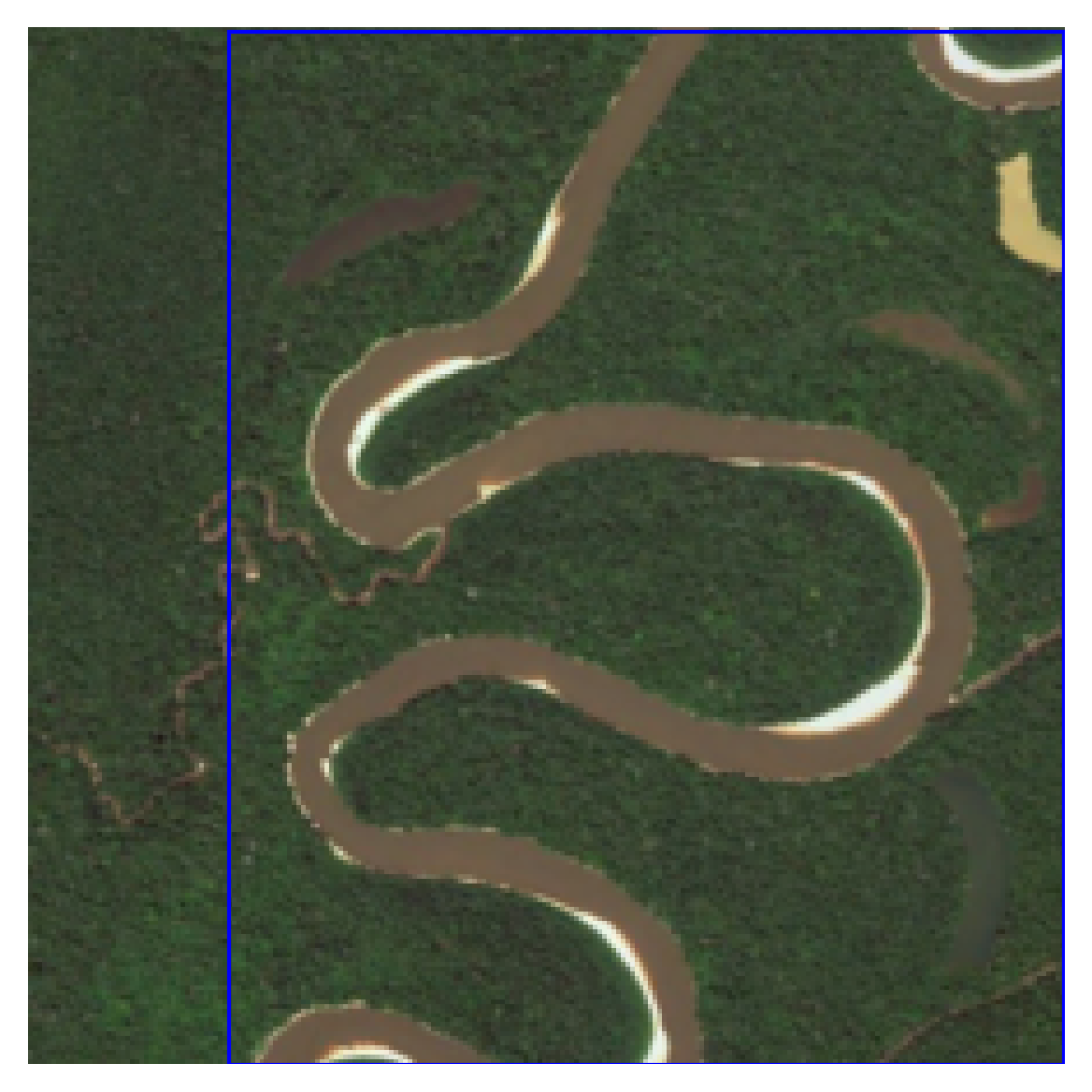}
            \caption*{Label: meander (1.00)\\Prediction: meander (1.00)}
        \end{subfigure}
    \end{minipage}
    \begin{minipage}{0.65\textwidth}
        \begin{subfigure}[]{\textwidth}
            \includegraphics[width=\textwidth]{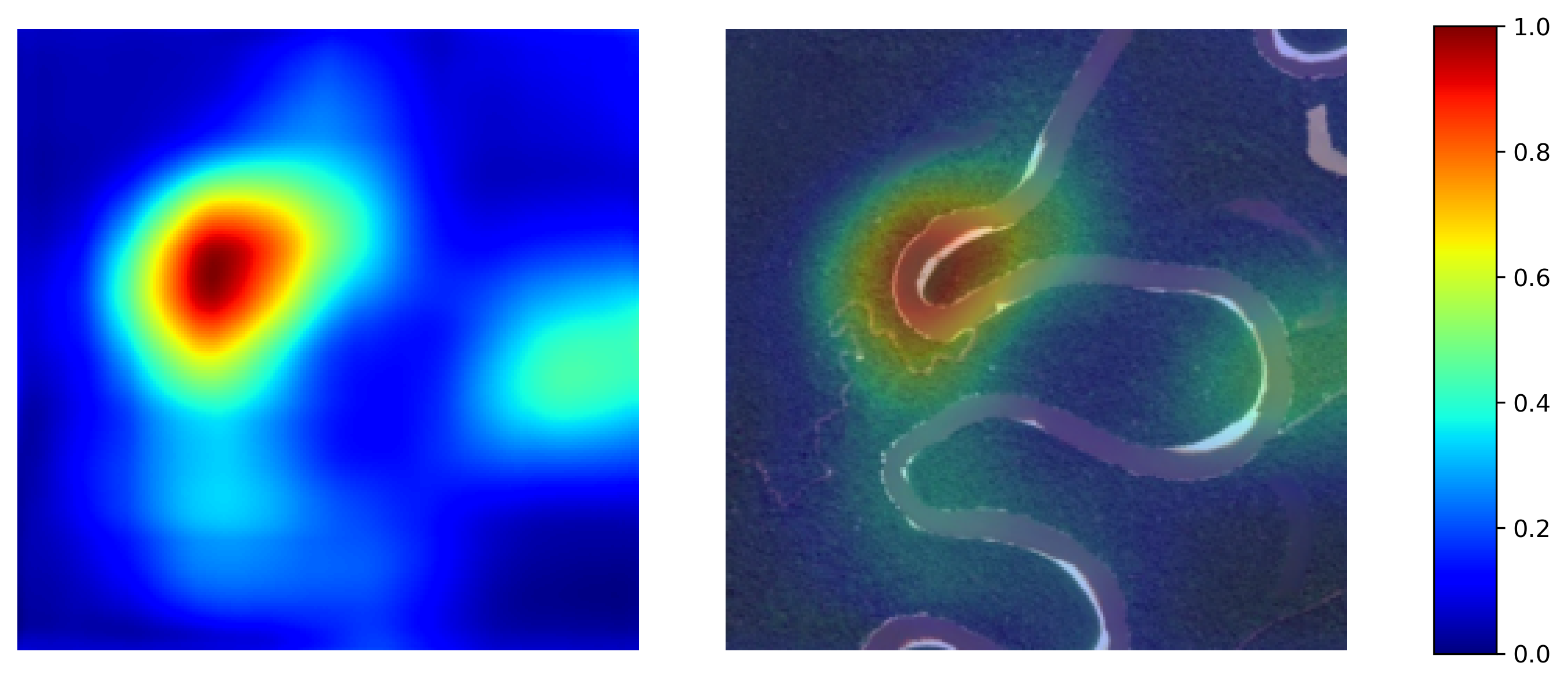}
            \caption{Occlusion}
            \label{fig_3a}
        \end{subfigure}
        \begin{subfigure}[]{\textwidth}
            \includegraphics[width=\textwidth]{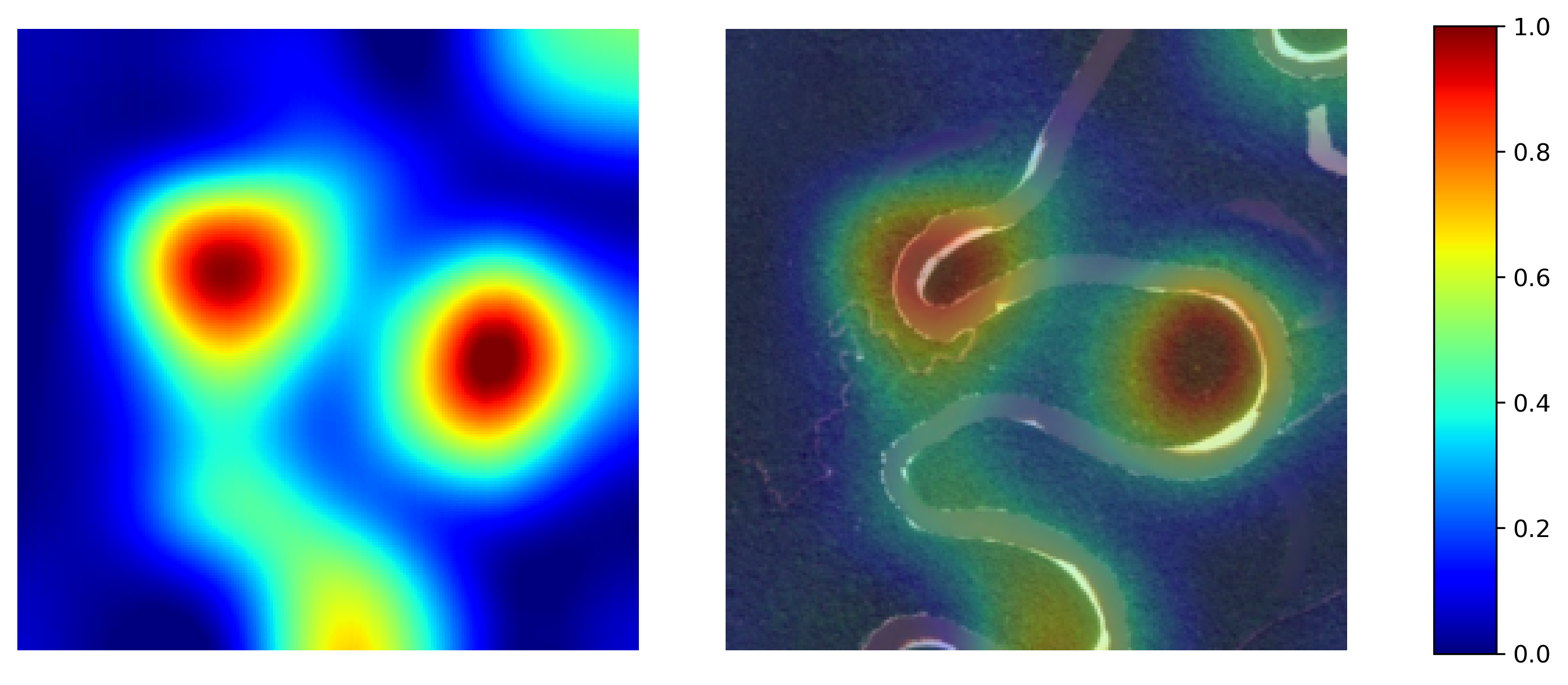}
            \caption{Grad-CAM}
            \label{fig_3b}
        \end{subfigure}
        \begin{subfigure}[]{\textwidth}
            \includegraphics[width=\textwidth]{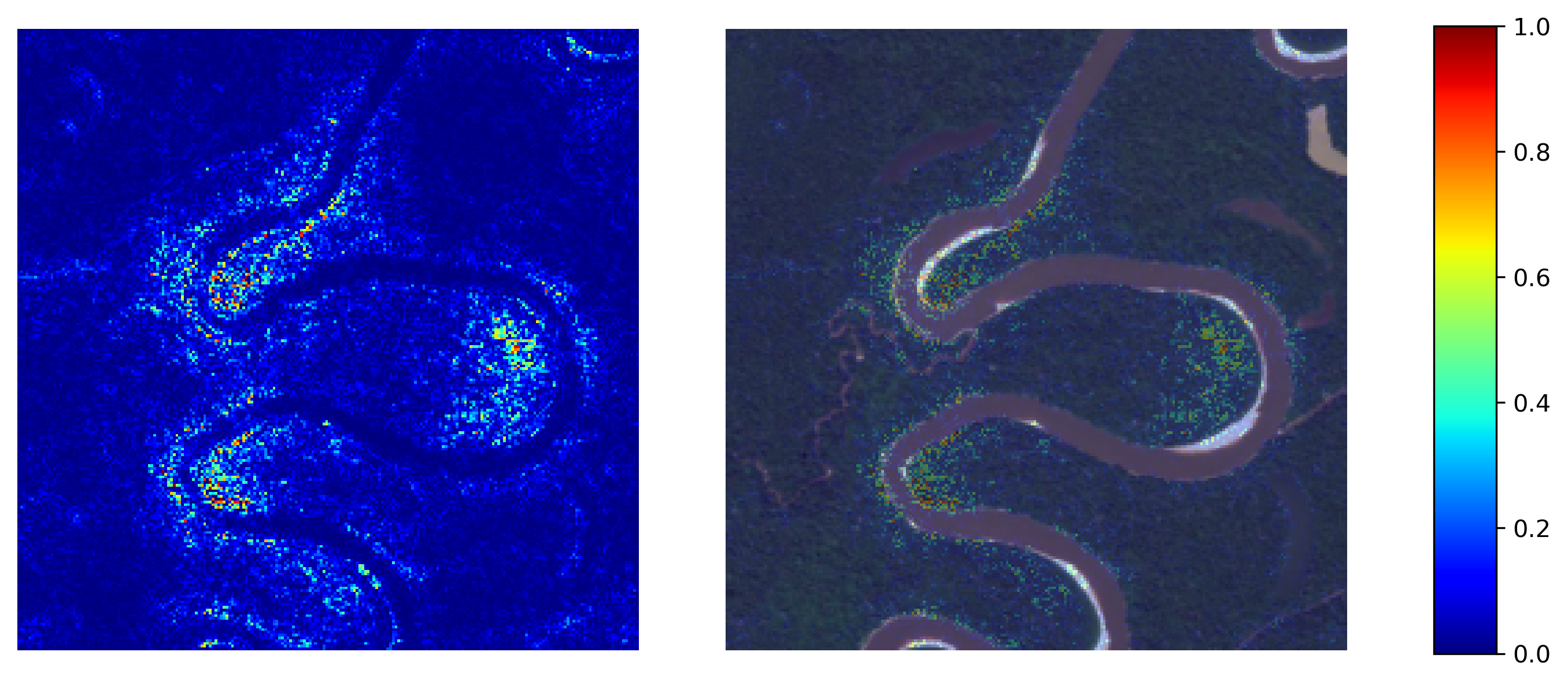}
            \caption{Integrated Gradients}
            \label{fig_3c}
        \end{subfigure}
    \end{minipage}
    \caption{Ability of the saliency maps to identify multiple prominent features of the same object. Figure layout is the same as that in Figure \ref{fig_2}.}
    \label{fig_3}
\end{figure}

Figure \ref{fig_3} shows the generated saliency maps for detecting meanders in images. A \textit{meander} is defined as “a river [that] flows back and forth across the landscape to form a series of sinuous curves” \citep{charlton2007fundamentals}. Hence, meanders are objects with multiple prominent features (i.e., curving parts). Similar to the detection of multiple objects (Figure \ref{fig_2}), the results related to the detection of multiple prominent features of the same object showed that Grad-CAM (Figure \ref{fig_3b}) and integrated gradients (Figure \ref{fig_3c}) could highlight multiple prominent features that characterize a meander. In contrast, the results from the occlusion method showed that the classification depends mainly on the detection of a single curve. Even though the occlusion method detected two curving parts of the meander (Figure \ref{fig_3a}), the curve near the center of the image received the most attention and garnered much more importance than did the curve near the right-side edge of the image (Figure \ref{fig_3a}). This is because, in the occlusion method, one prominent part of the object dominantly influences the result and saturates the output probability. 

\subsection{Ability to accurately capture object shape}
\begin{figure}
    \captionsetup[subfigure]{justification=centering, skip=-2pt}
    \centering
    \begin{minipage}{0.3\textwidth}
        \begin{subfigure}[]{\textwidth}
            \includegraphics[width=\textwidth]{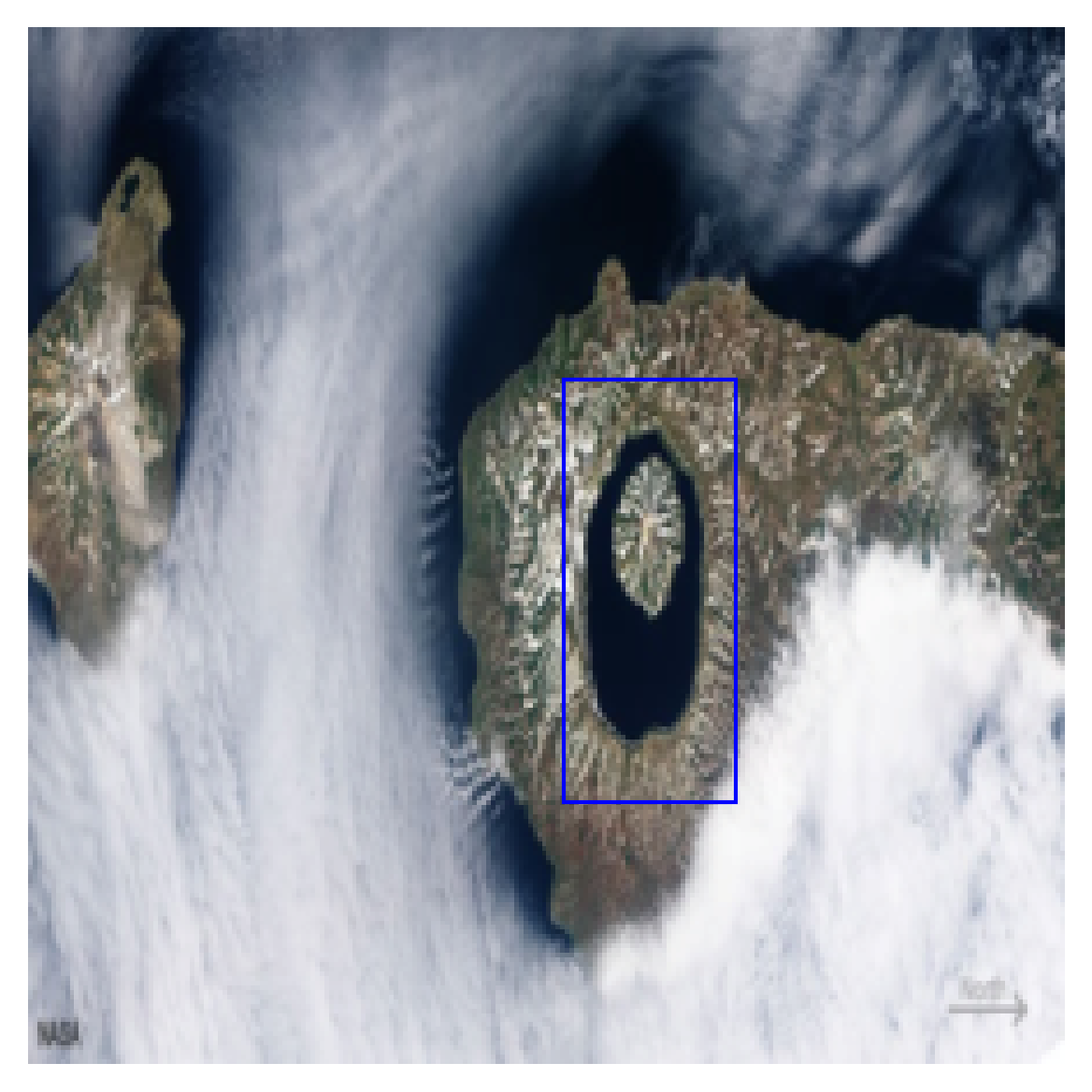}
            \caption*{Label: crater (1.00)\\Prediction: crater (1.00)}
        \end{subfigure}
    \end{minipage}
    \begin{minipage}{0.65\textwidth}
        \begin{subfigure}[]{\textwidth}
            \includegraphics[width=\textwidth]{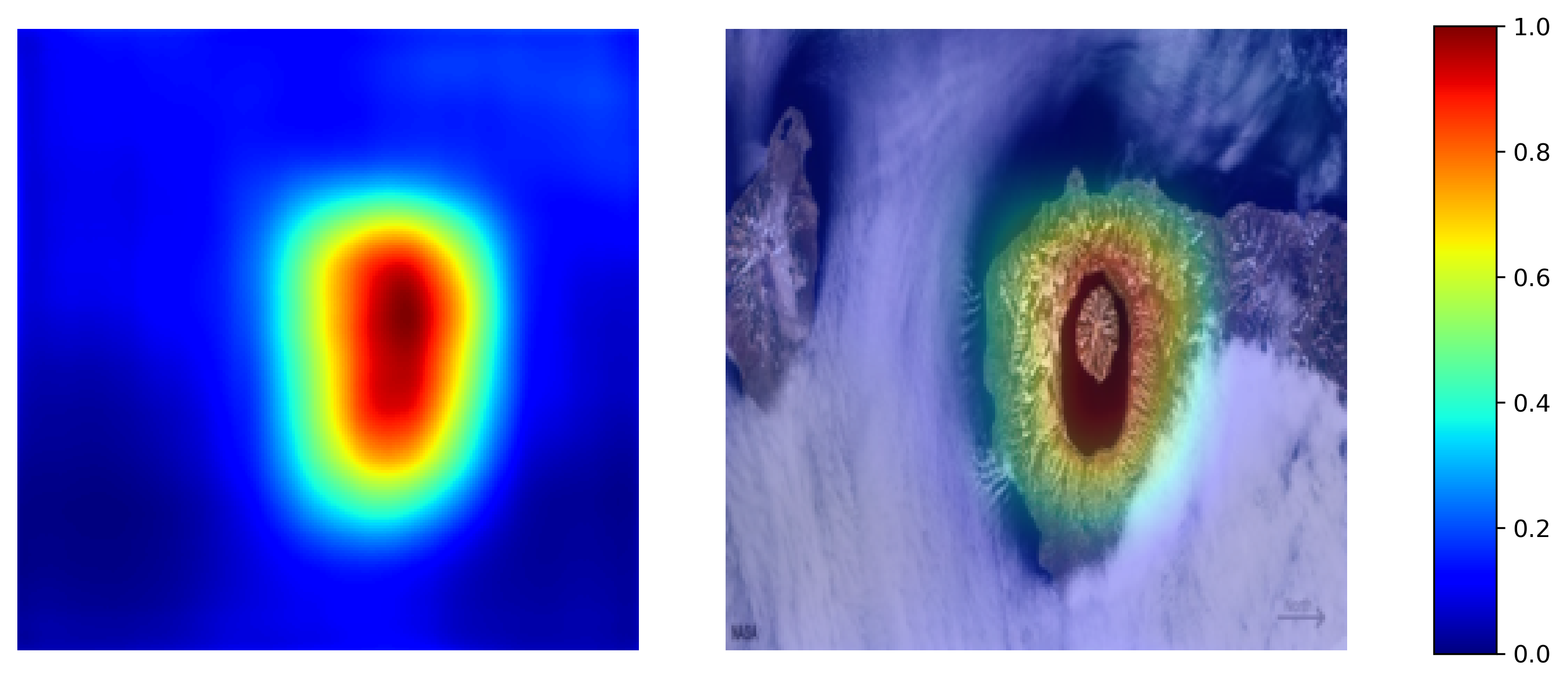}
            \caption{Occlusion}
            \label{fig_4a}
        \end{subfigure}
        \begin{subfigure}[]{\textwidth}
            \includegraphics[width=\textwidth]{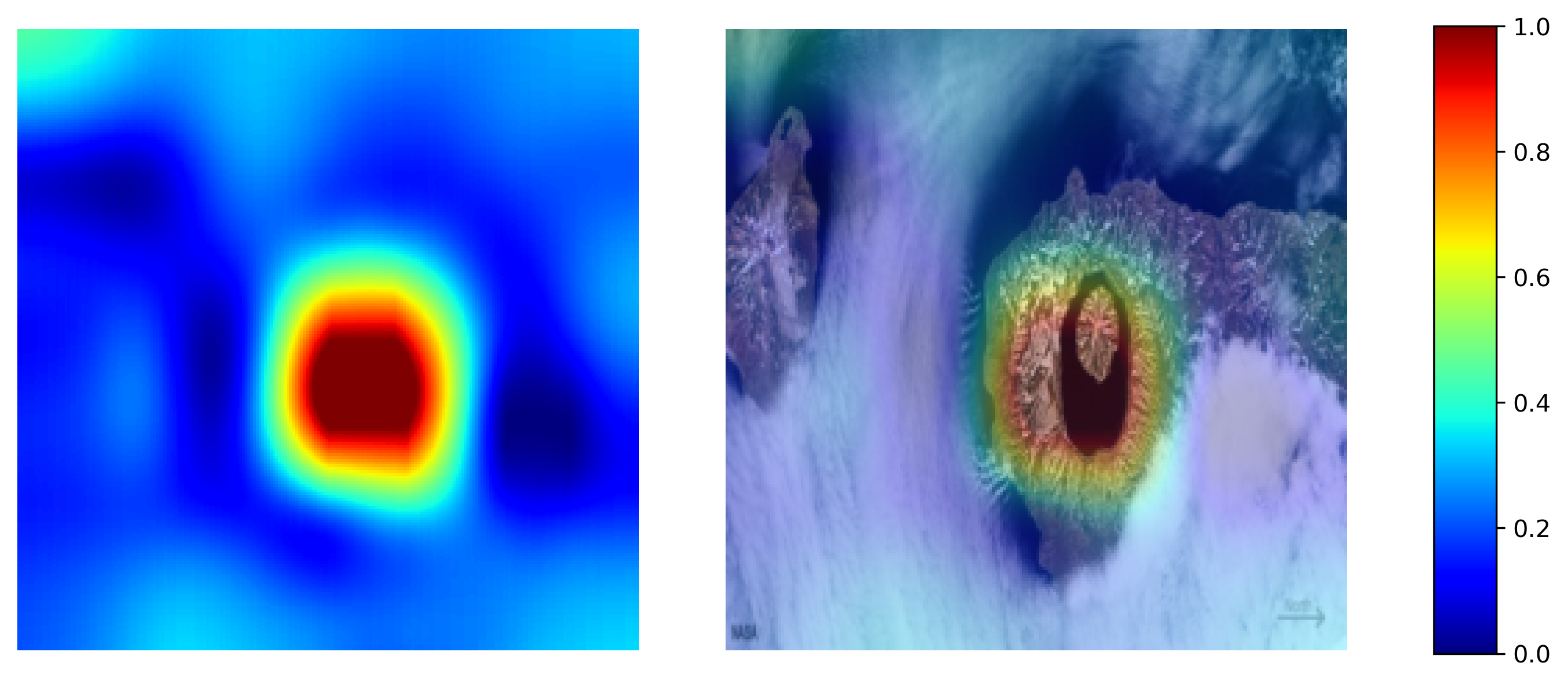}
            \caption{Grad-CAM}
            \label{fig_4b}
        \end{subfigure}
        \begin{subfigure}[]{\textwidth}
            \includegraphics[width=\textwidth]{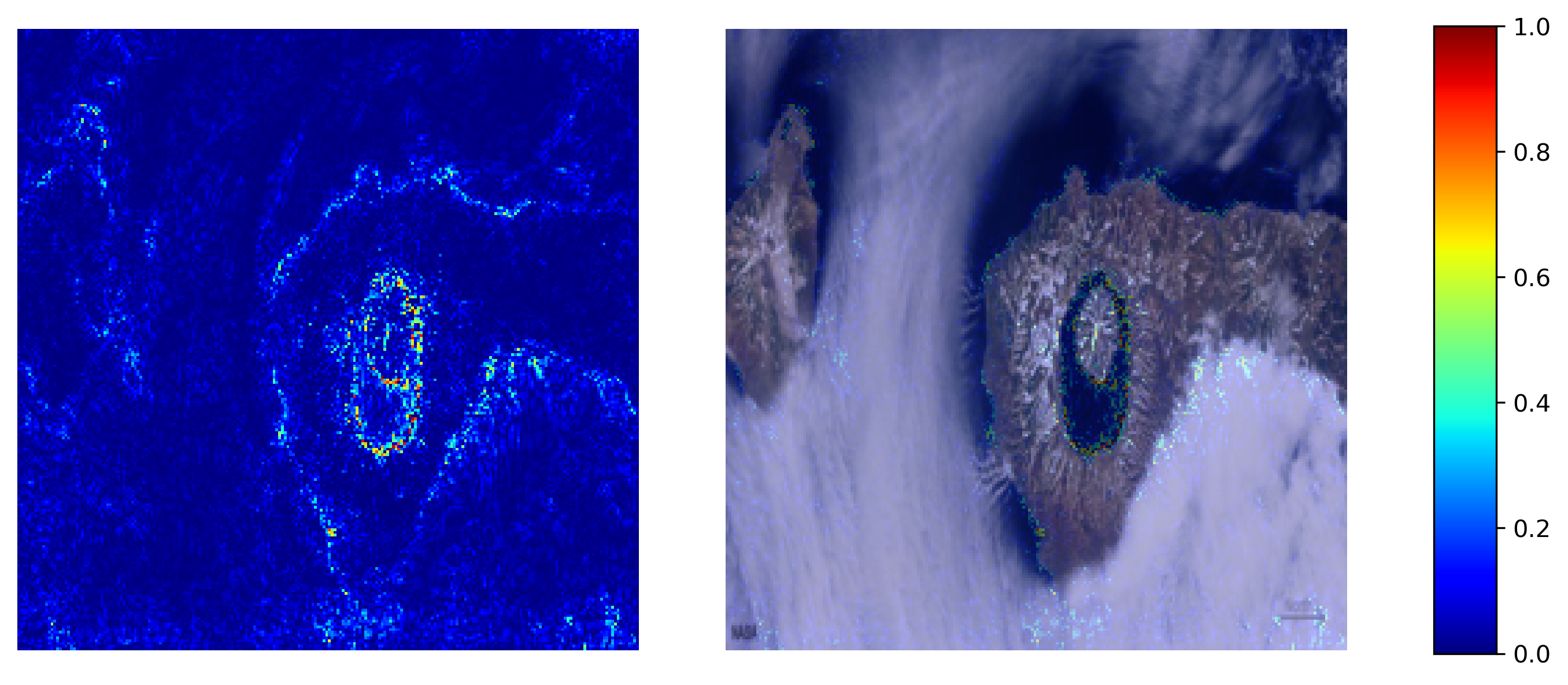}
            \caption{Integrated Gradients}
            \label{fig_4c}
        \end{subfigure}
    \end{minipage}
    \caption{Accuracy of the saliency maps in highlighting the shape of the target objects. Figure layout is the same as that in Figure \ref{fig_2}.}
    \label{fig_4}
\end{figure}

Figure \ref{fig_4} shows a comparison of the different methods in correctly highlighting the shape of an oval shaped crater. Both the occlusion (Figure \ref{fig_4a}) and the integrated gradients (Figure \ref{fig_4c}) methods can generate highlights that match the actual shapes of the target (i.e., a crater with an oval shape, shown in Figure \ref{fig_4a}) because they both calculate the importance of image regions at the pixel-level of the input image. As pixel values near the edges of the objects often have shape changes, the pixel-based saliency map generation process can relatively accurately capture the border pixels and, hence, the shape of the target. Grad-CAM, however, may generate mismatching shapes between the highlighted image areas and the target shape in its visualization. As shown in Figure \ref{fig_4b}, a circular region was generated by Grad-CAM when interpreting the oval-shaped crater. This is because Grad-CAM combines feature maps from the last convolutional layer -- which has a much lower resolution than does the original image -- to generate the saliency map, the resolution of which is therefore lower than that generated by the other two methods. This coarse saliency map is often upsampled to the same size as the input image to gain a better visual effect. However, the upsampling may cause information loss, and some subtle differences (e.g., between a circular shaped area vs. an oval-shaped area) at a coarse resolution may become more substantial after upsampling, resulting in location and shape mismatches. 

\subsection{Ability to correctly highlight important image regions}
\begin{figure}
    \captionsetup[subfigure]{justification=centering, skip=-2pt}
    \centering
    \begin{minipage}{0.3\textwidth}
        \begin{subfigure}[]{\textwidth}
            \includegraphics[width=\textwidth]{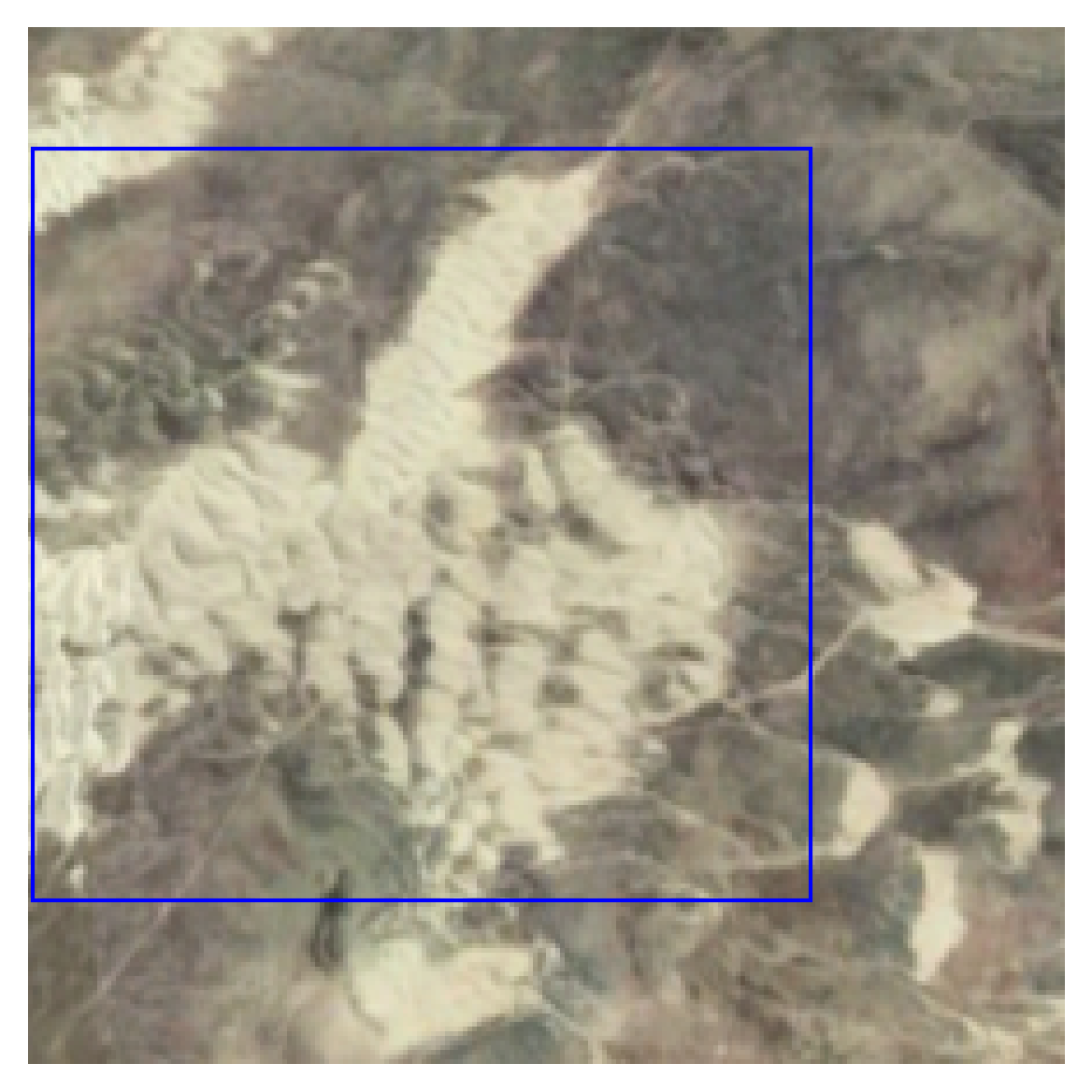}
            \caption*{Label: dunes (1.00)\\Prediction: dunes (1.00)}
        \end{subfigure}
    \end{minipage}
    \begin{minipage}{0.65\textwidth}
        \begin{subfigure}[]{\textwidth}
            \includegraphics[width=\textwidth]{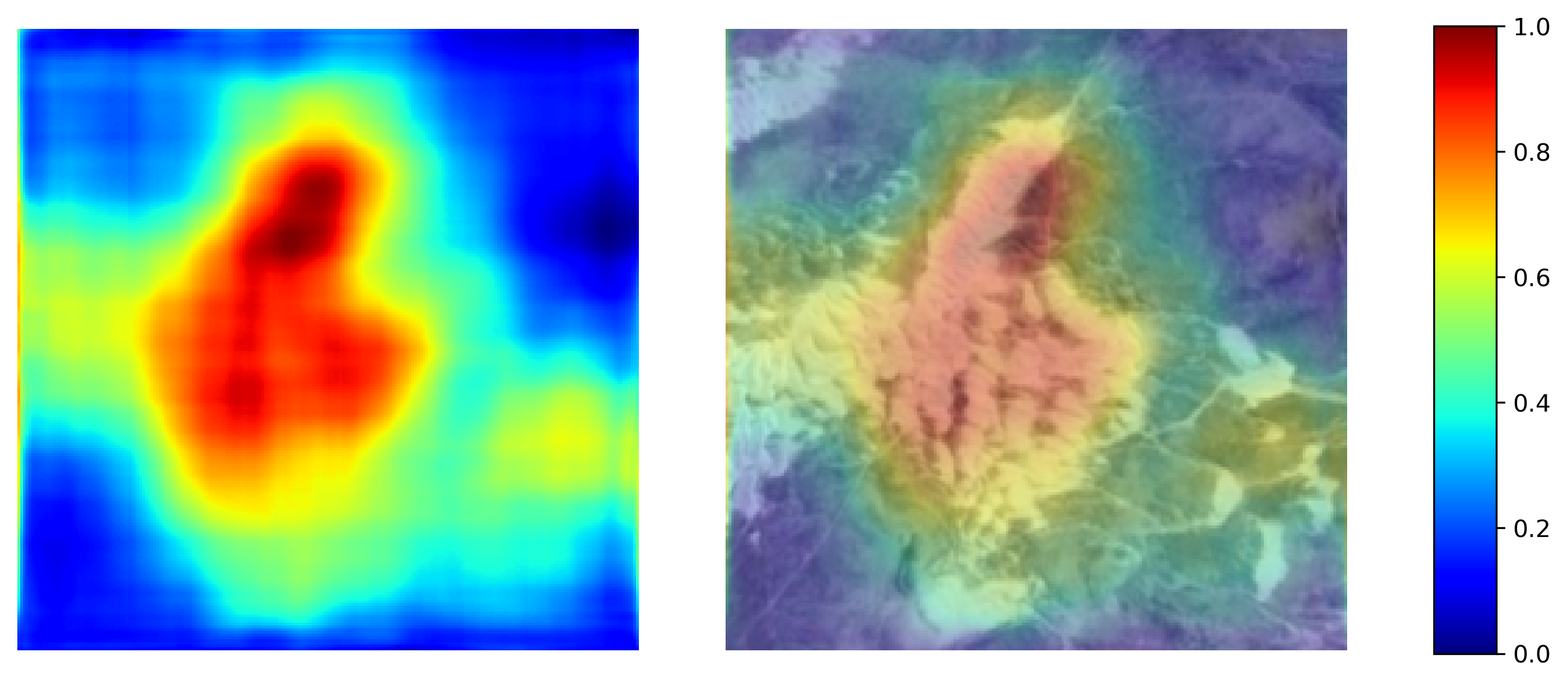}
            \caption{Occlusion}
            \label{fig_5a}
        \end{subfigure}
        \begin{subfigure}[]{\textwidth}
            \includegraphics[width=\textwidth]{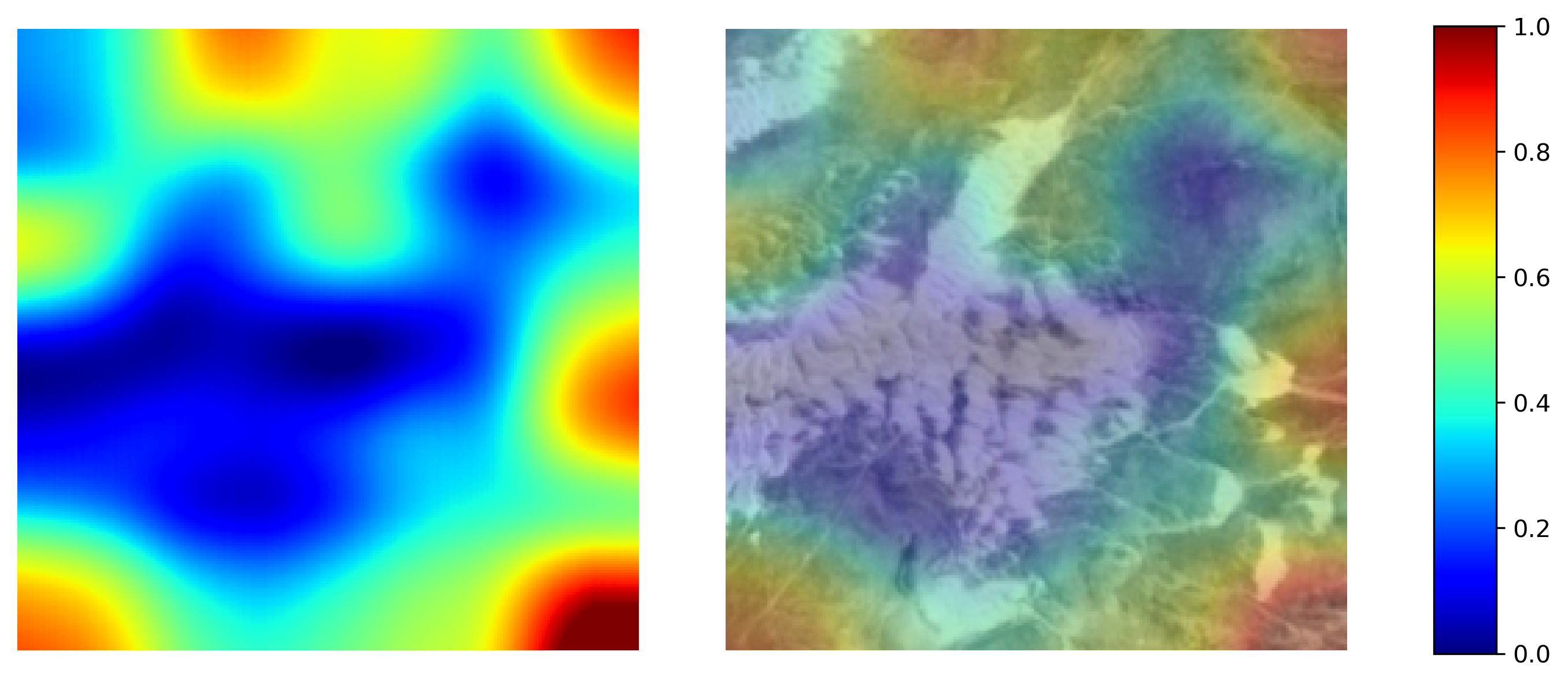}
            \caption{Grad-CAM}
            \label{fig_5b}
        \end{subfigure}
        \begin{subfigure}[]{\textwidth}
            \includegraphics[width=\textwidth]{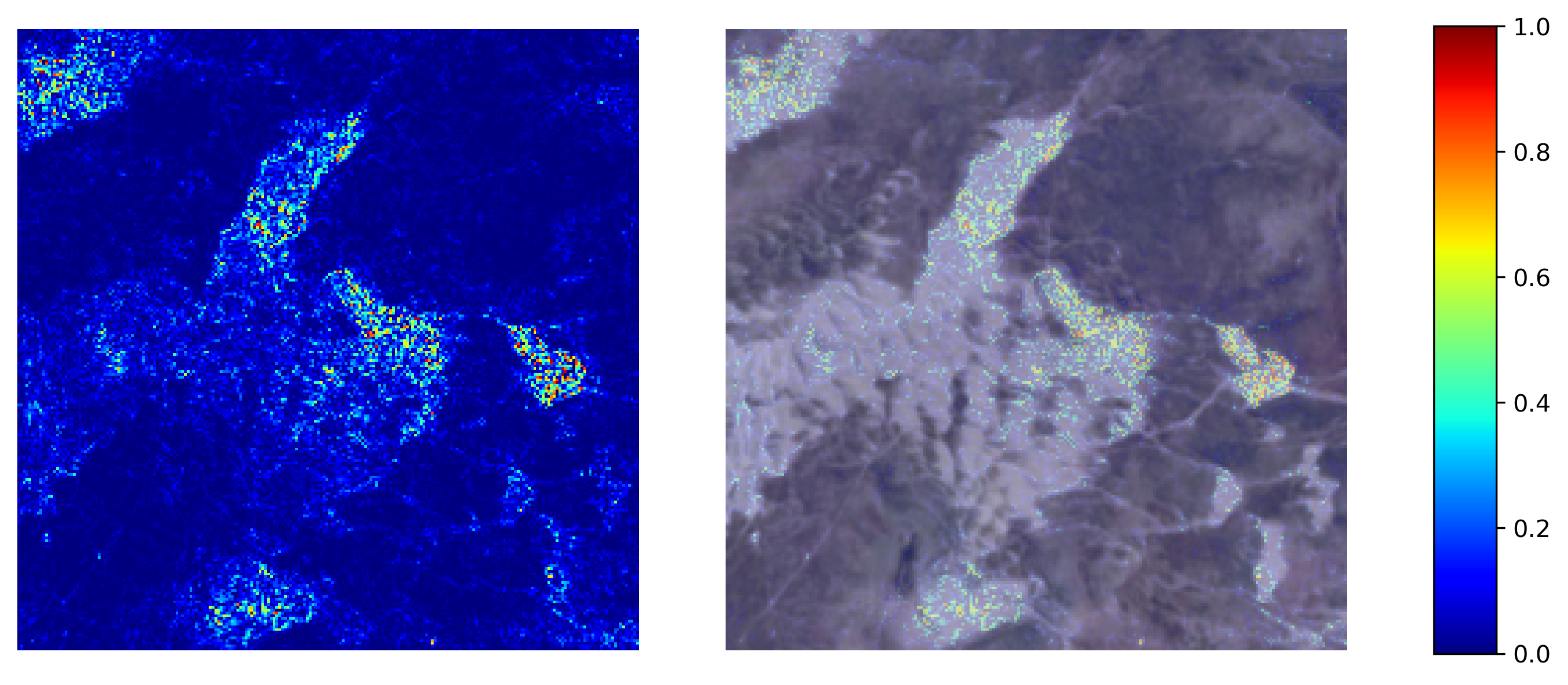}
            \caption{Integrated Gradients}
            \label{fig_5c}
        \end{subfigure}
    \end{minipage}
    \caption{Accuracy of the saliency maps in highlighting important image areas containing the target objects. Figure layout is the same as that in Figure \ref{fig_2}.}
    \label{fig_5}
\end{figure}

Figure \ref{fig_5} shows a comparison of the three methods in terms of their ability to correctly highlight important image regions in the resultant saliency maps. For this test, the object to detect was sand dunes. Results indicate that both the occlusion and the integrated gradients methods were able to highlight the targets correctly, and their results were similar. However, Grad-CAM highlighted surrounding areas of the target instead of the target itself. Analyzing the algorithm behaviors, we note that the large differences between the results generated by Grad-CAM and the other two methods may be caused by the issue of gradient discontinuity, a non-linear function used in the Grad-CAM model. As does ReLU, it introduces discontinuous gradients when they are non-differentiable at some locations of the function curves. Grad-CAM computes partial derivatives of the classification score with respect to each pixel at the last convolutional layer, and the discontinuity may transfer to some artifacts in the saliency map. Unlike Grad-CAM, the occlusion and integrated gradients methods do not suffer from this issue. The occlusion method takes an image as the input and observes the output probability change in the forward pass, while the integrated gradients method computes and integrates all gradients from the baseline image to the input image to avoid the effect of discontinuous gradients. The results in Figure \ref{fig_5} further illustrate the advantages of using a joint analysis approach to avoid the possible pitfalls of using a single method. 

\subsection{Ability to generate clear areas of attention due to the existence of multiple high-contrast areas}
\begin{figure}
    \captionsetup[subfigure]{justification=centering, skip=-2pt}
    \centering
    \begin{minipage}{0.3\textwidth}
        \begin{subfigure}[]{\textwidth}
            \includegraphics[width=\textwidth]{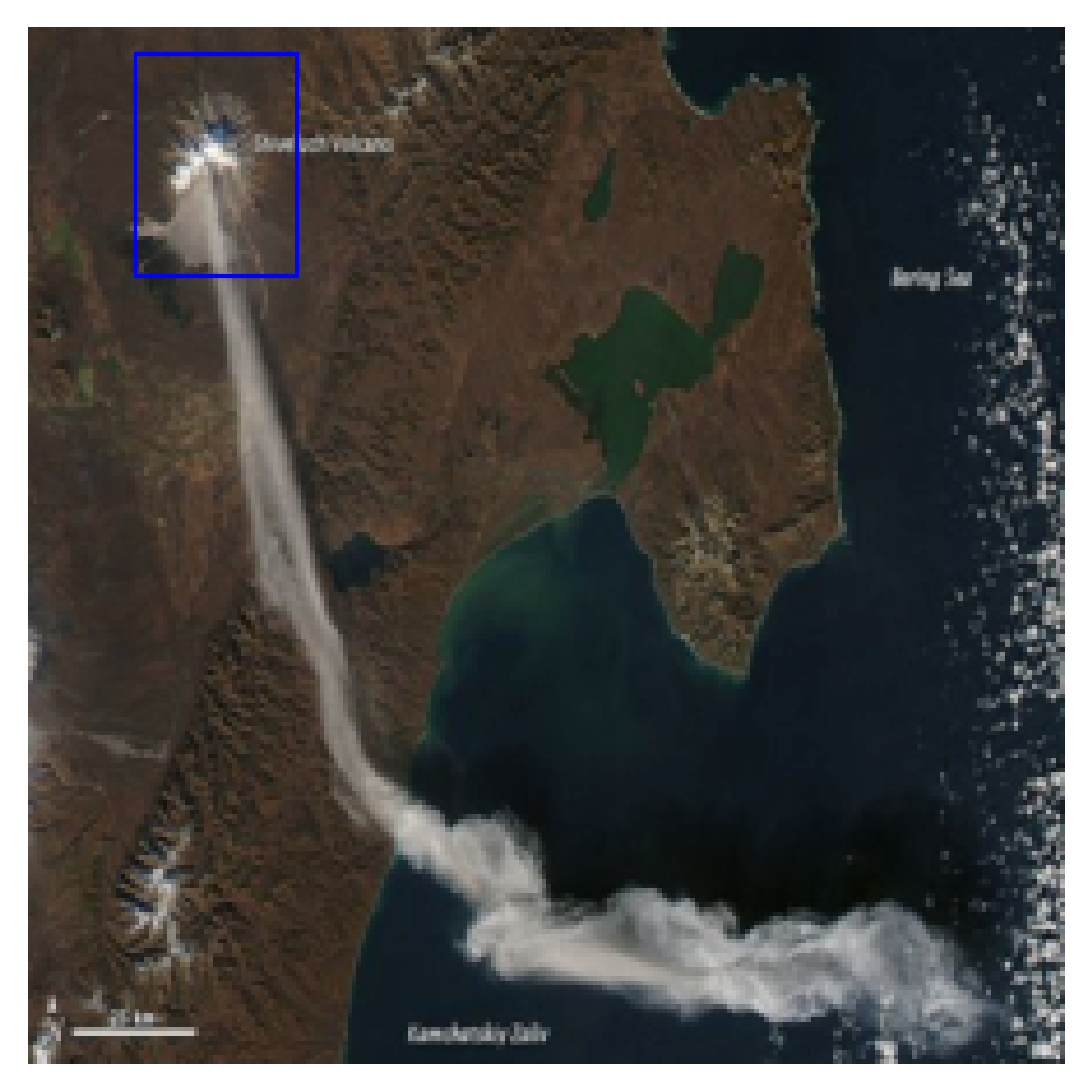}
            \caption*{Label: volcano (1.00)\\Prediction: volcano (1.00)}
        \end{subfigure}
    \end{minipage}
    \begin{minipage}{0.65\textwidth}
        \begin{subfigure}[]{\textwidth}
            \includegraphics[width=\textwidth]{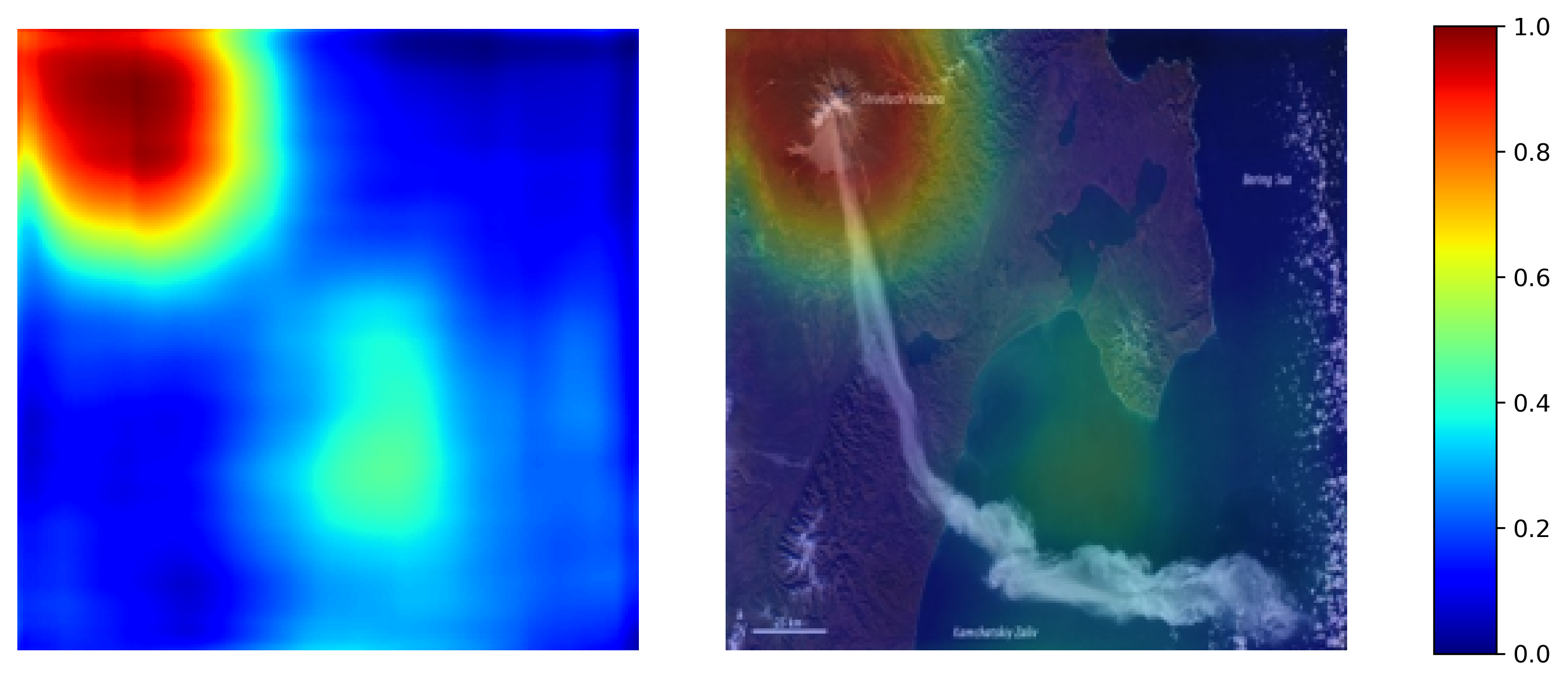}
            \caption{Occlusion}
            \label{fig_6a}
        \end{subfigure}
        \begin{subfigure}[]{\textwidth}
            \includegraphics[width=\textwidth]{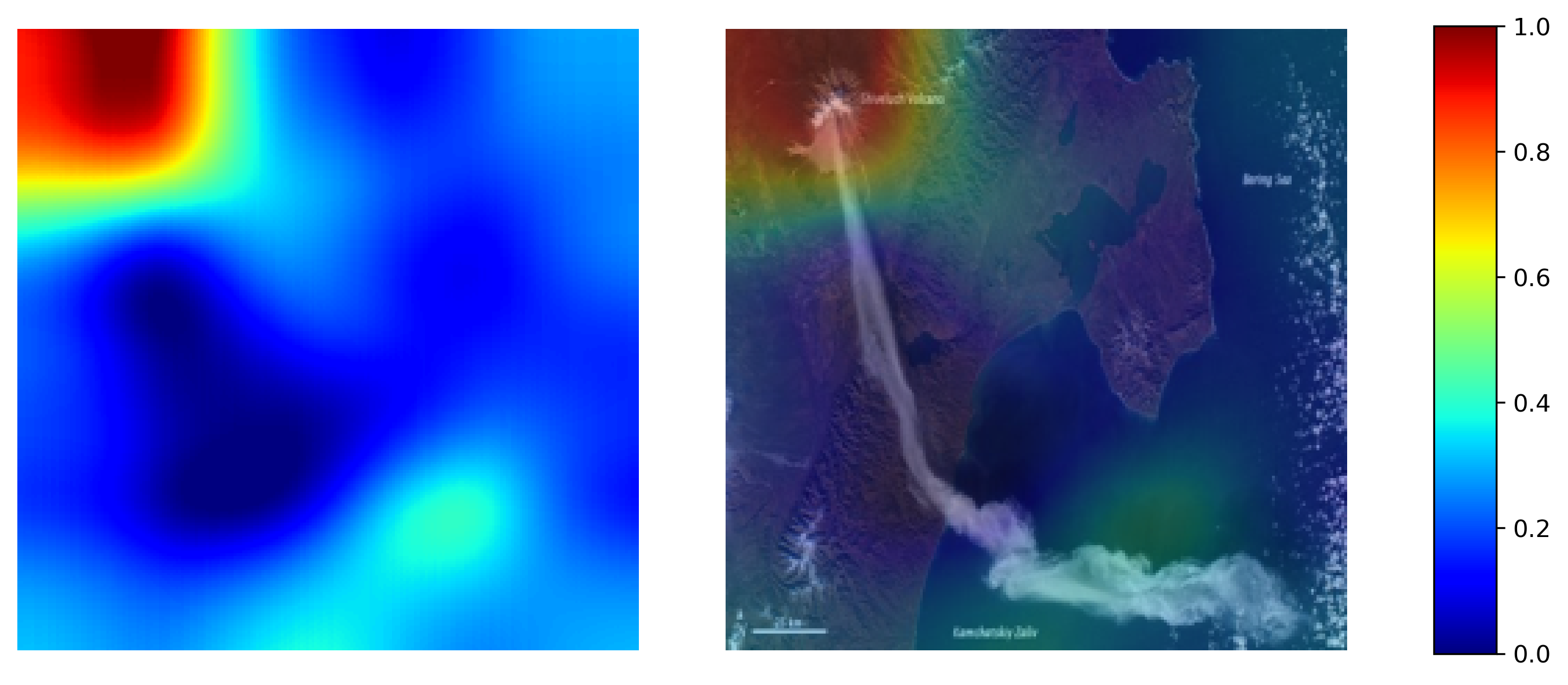}
            \caption{Grad-CAM}
            \label{fig_6b}
        \end{subfigure}
        \begin{subfigure}[]{\textwidth}
            \includegraphics[width=\textwidth]{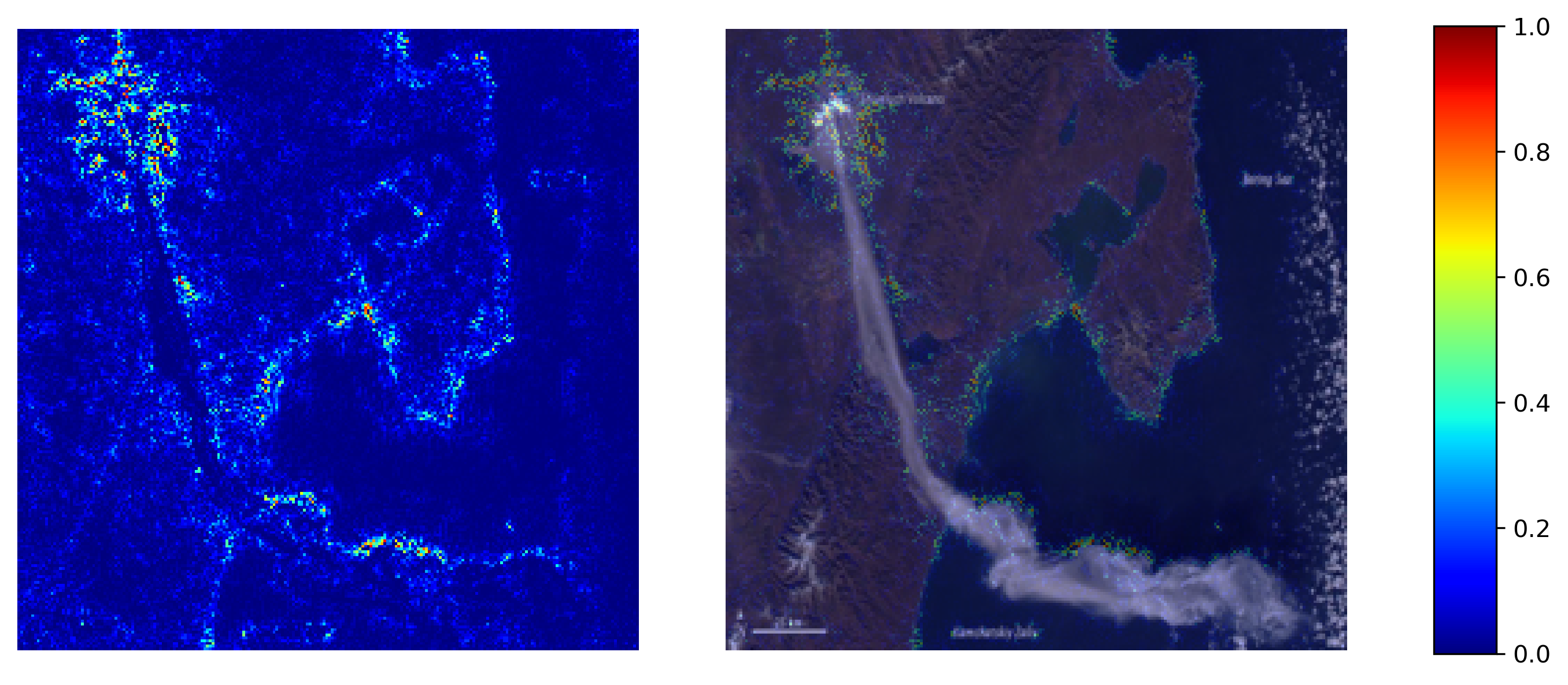}
            \caption{Integrated Gradients}
            \label{fig_6c}
        \end{subfigure}
    \end{minipage}
    \caption{Comparison of the saliency maps in generating clear attention when multiple high-contrast areas exist. Figure layout is the same as that in Figure \ref{fig_2}.}
    \label{fig_6}
\end{figure}

Figure \ref{fig_6} shows the saliency maps generated by different methods when there are multiple high-contrast areas presented in an image. In this figure, an erupting volcano was the target feature to detect. In addition to the volcano, the image contained erupted gases forming a unique linear shape, which had a high level of contrast with the background. Meanwhile, the coastline in the image also presented a high level of contrast, forming the division between the water and the land. Because the integrated gradients approach uses per-pixel computation to generate pixel-wise highlights, it can more easily distinguish edges and even thin linear features, such as ridges or shorelines. For the case presented in Figure \ref{fig_6}, the integrated gradients method highlighted multiple areas with such characteristics (Figure \ref{fig_6c}). However, this result makes it difficult to examine which are the most distinctive image regions that guided the model to detect the target of interest (i.e., volcano). The occlusion and Grad-CAM approaches, in comparison, can more confidently and clearly highlight the most important image regions that the model relies on to make a correct prediction (i.e., the mouth of the volcano). This issue of the integrated gradients approach is also found in processing images when there is a low level of contrast between the target and the background, such as in the detection of an iceberg tongue in an image of a glacier. 

\subsection{Ability to generate clear attention due to low contrast between the foreground and background}
\begin{figure}
    \captionsetup[subfigure]{justification=centering, skip=-2pt}
    \centering
    \begin{minipage}{0.3\textwidth}
        \begin{subfigure}[]{\textwidth}
            \includegraphics[width=\textwidth]{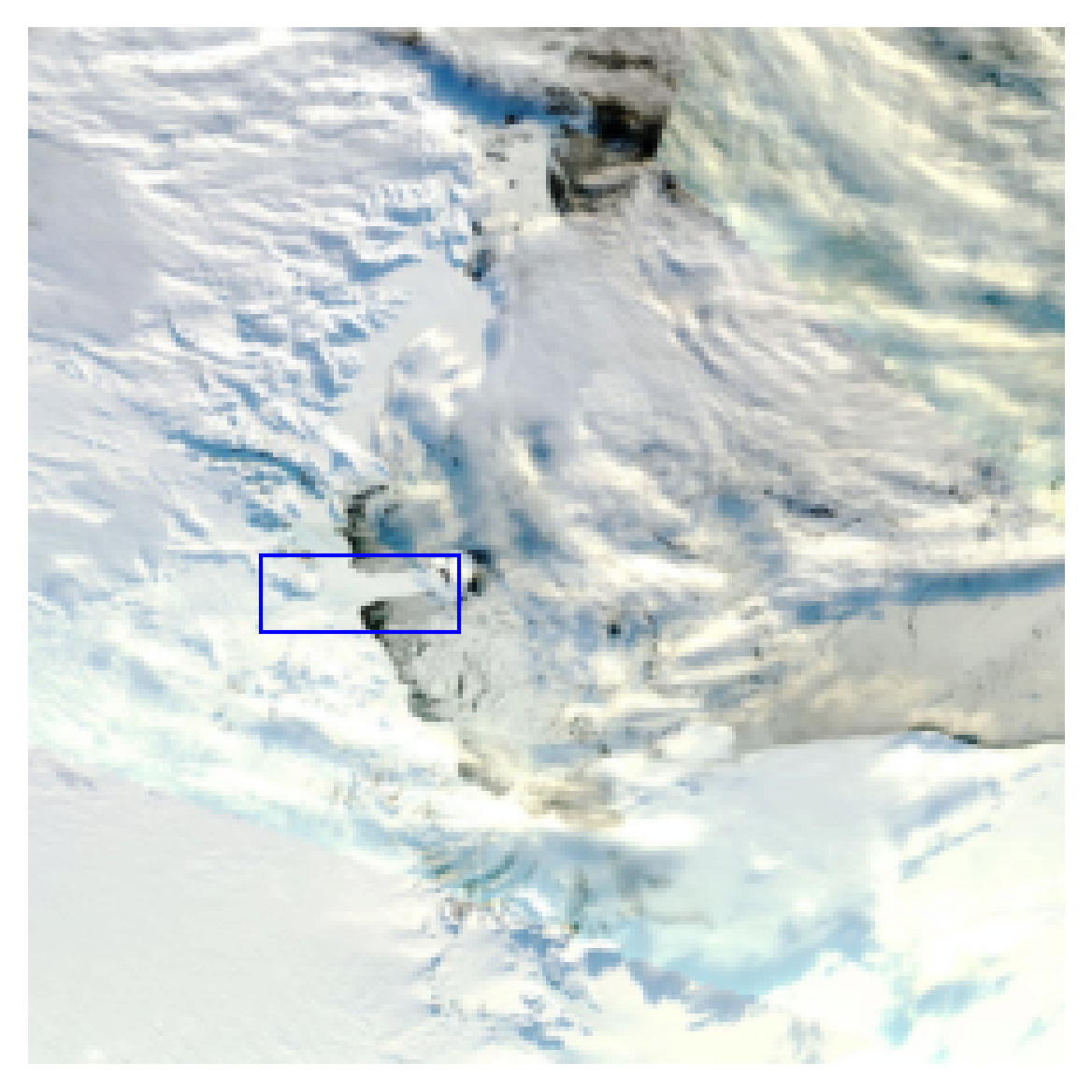}
            \caption*{Label: iceberg tongue (1.00)\\Prediction: iceberg tongue (1.00)}
        \end{subfigure}
    \end{minipage}
    \begin{minipage}{0.65\textwidth}
        \begin{subfigure}[]{\textwidth}
            \includegraphics[width=\textwidth]{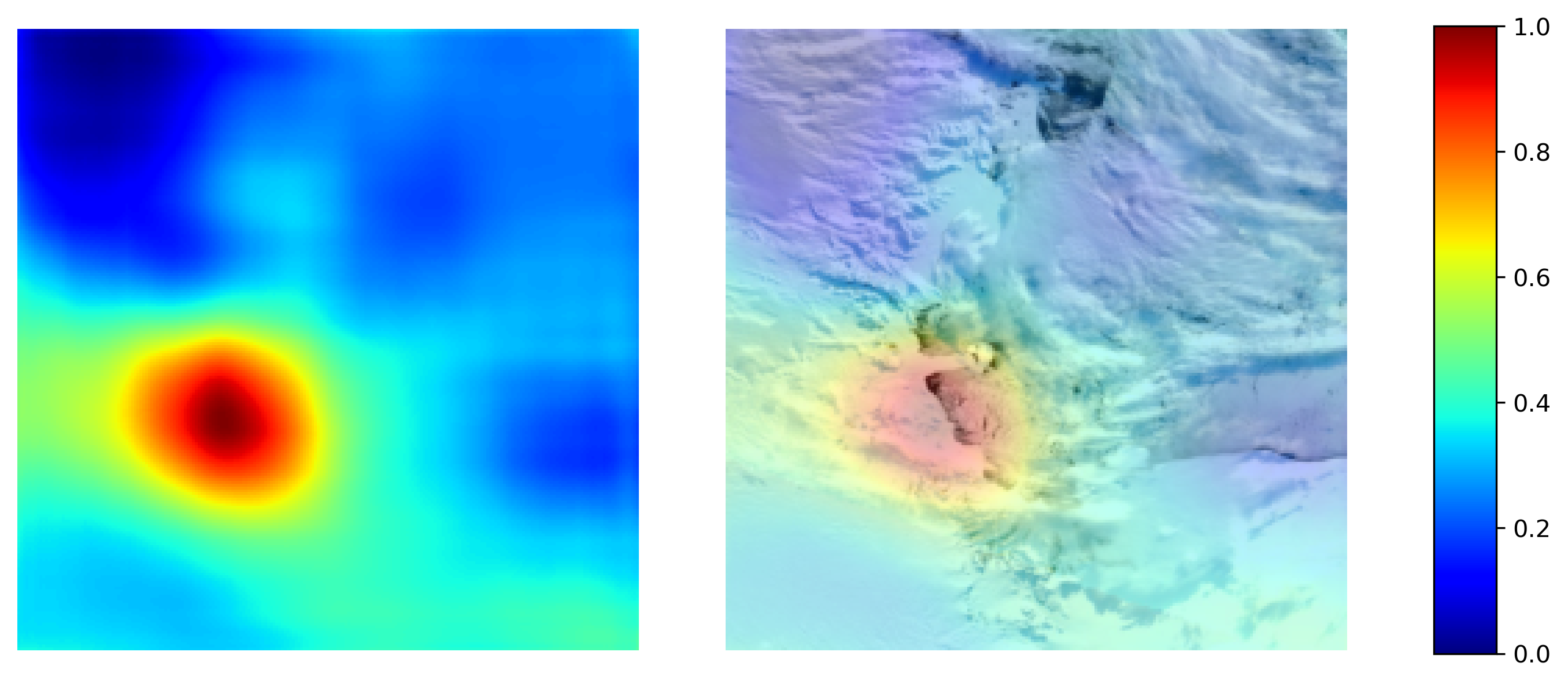}
            \caption{Occlusion}
            \label{fig_7a}
        \end{subfigure}
        \begin{subfigure}[]{\textwidth}
            \includegraphics[width=\textwidth]{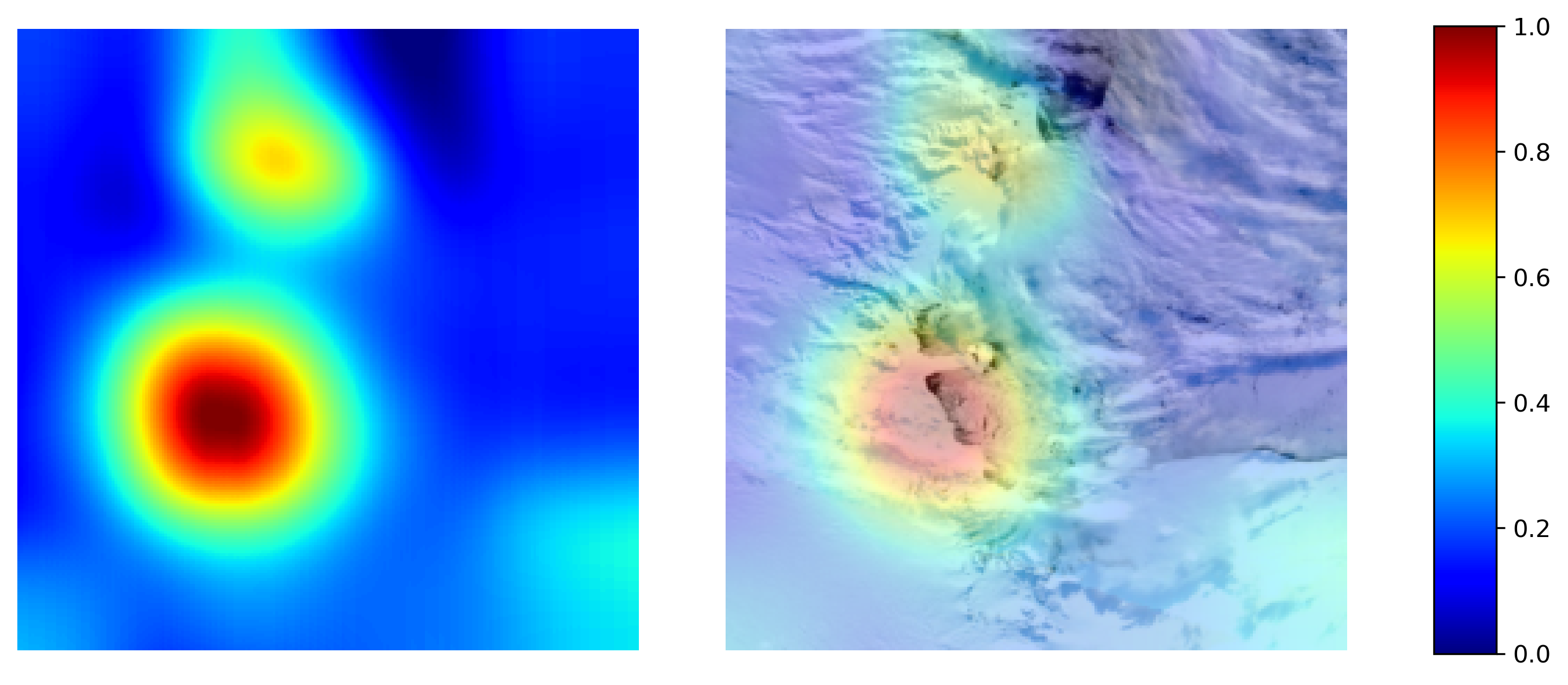}
            \caption{Grad-CAM}
            \label{fig_7b}
        \end{subfigure}
        \begin{subfigure}[]{\textwidth}
            \includegraphics[width=\textwidth]{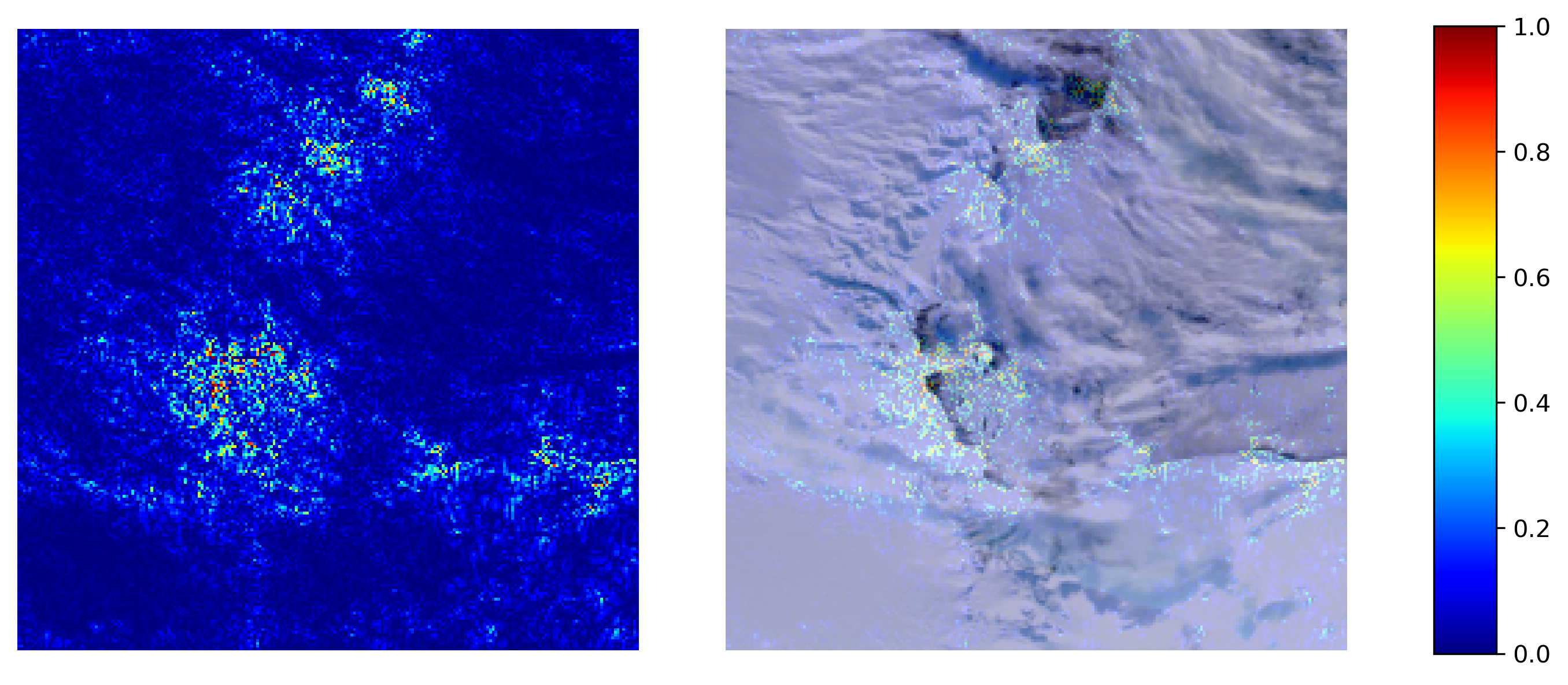}
            \caption{Integrated Gradients}
            \label{fig_7c}
        \end{subfigure}
    \end{minipage}
    \caption{Comparison of the saliency maps in generating clear attention when the contrast between the foreground and background is low. Figure layout is the same as that in Figure \ref{fig_2}.}
    \label{fig_7}
\end{figure}

Figure \ref{fig_7} further illustrates the contrast-related issues of the integrated gradients method, which results in unclear patterns when the contrast between the foreground (i.e., the target) and the background of an image is low. As discussed, the integrated gradients approach can generate a saliency map at the pixel level, providing a fine-grained view of the important pixels and image subareas identified by a model when making decisions. However, under some conditions, the highlighted pixels could spread across the image, making the explanation of the model's decision process difficult to achieve. For instance, Figure \ref{fig_7} presents an iceberg tongue feature within a glacier region. Because an iceberg tongue is often a part of a glacier, and they are both white, there tends to be a low level of contrast between the two features. The integrated gradients approach generates pixel-level highlights based on the existence of shades, which may be randomly distributed in the image. Accordingly, even though the iceberg tongue area has been highlighted in the result (Figure \ref{fig_7c}), other areas are also highlighted, making it difficult to understand and explain the model's learning process. In comparison, the occlusion and the Grad-CAM methods create saliency maps with region-based patterns containing less noise, providing a clearer view on the important feature and contextual information that helps the model with its prediction. 

\section{Research Findings}
\subsection{A summary analysis of CNN model explanation methods}
Table \ref{tbl_1} summarizes the capabilities of the model explanation methods. We also generate the statistics (Table \ref{tbl_2}) based on the entire training set by counting the number of saliency maps generated by these methods that can meet an explanation goal. The cells in Table \ref{tbl_1} are labeled as “Often Yes” when a method (e.g., Grad-CAM) meets the goal in most cases; they are labeled as “Often No” when a method does not meet an explanation goal in most testing cases. A 100\% “Yes” case does not exist due to the inherent uncertainties in the model explanation process for calculating attributed features and the inconsistency between model perception and human perception.  

In summary, Grad-CAM can highlight multiples of the same type of object and multiple prominent features of a single object. It can also generate clear attention by highlighting important image regions when there are multiple high-contrast areas in an image or when the contrast between the foreground (i.e., the target) and background is low. However, it sometimes cannot accurately reveal the object shape due to the saliency map’s low resolution. Comparatively, the occlusion and integrated gradient approaches can better reveal an object’s actual shape as they generate the saliency map at the input images’ resolutions. The pixel-based computation adopted in the integrated gradient approach makes it capable to identify multiple objects or multiple prominent features of an object. However, when an image has multiple high-contrast areas, this method may highlight all of them, generating noise in the saliency map. Although the occlusion approach avoids highlighting multiple high-contrast areas if they belong to different object classes, the gradient saturation issue makes it favor only one of multiple objects of the same class or one prominent feature of an object. 

The Grad-CAM method has the advantage of greater computational efficiency because it requires only a single forward and partial backward computation per image. In contrast, the occlusion and the integrated gradients methods need multiple forward and backward calculations. Their computation efforts depend on corresponding hyperparameters, which incur a trade-off between saliency map quality and computation time. Compared to methods such as CAM \citep{zhou2016learning}, none of the methods studied in this paper require network architecture changes; therefore, there is no need to retrain the deep learning models.

\begin{table}
\tbl{Summary analysis of the capabilities of the CNN model explanation methods.}
{\begin{tabular}{p{0.4\textwidth}ccc} \toprule
 Model Explanation Goals & Occlusion Approach & Grad-CAM & Integrated Gradients \\ \midrule
 Ability to identify multiple objects of the same type & Often No & Often Yes & Often Yes \\
 Ability to identify multiple prominent features of a single object & Often No & Often Yes & Often Yes \\
 Ability to accurately detect object shape & Often Yes & Sometimes No & Often Yes \\
 Ability to correctly highlight important image regions	& Often Yes & Sometimes No & Often Yes \\
 Ability to generate clear attentions for an image with multiple high-contrast areas & Often Yes & Often Yes & Sometimes No \\
 Ability to generate clear attentions for an image when the contrast between the target and the background is low & Often Yes & Often Yes & Sometimes No \\
 Computational efficiency & Low & High & Low \\
 Need to retrain the model & No & No & No \\ \bottomrule
\end{tabular}}
\label{tbl_1}
\end{table}

\begin{table}
\tbl{Statistics on percentage of saliency maps generated by different methods that can meet an explanation goal. ``Total'' means number of images in the dataset that supports the corresponding criterion listed in each row. 
}
{\begin{tabular}{p{0.4\textwidth}cccc} \toprule
 Model Explanation Goals & Occlusion Approach  & Grad-CAM  & Integrated Gradients  & Total \\ \midrule
 Ability to identify multiple objects of the same type & 30\% & 71\% & 79\% & 73 \\
 Ability to identify multiple prominent features of a single object & 31\% & 81\% & 89\% & 214 \\
 Ability to accurately detect object shape & 76\% & 44\% & 79\% & 826 \\
 Ability to correctly highlight important image regions	& 81\% & 58\% & 87\% & 826 \\
 Ability to generate clear attentions for an image with multiple high-contrast areas & 86\% & 87\% & 53\% & 207 \\
 Ability to generate clear attentions for an image when the contrast between the target and the background is low & 82\% & 78\% & 50\% & 152 \\
 \\ \bottomrule
\end{tabular}}
\label{tbl_2}
\end{table}

\subsection{Results generalizability}
To demonstrate the generalizability of our results, we adopted another AI-ready natural feature dataset—GeoNat v1.0 \citep{arundel2020geonat}. This dataset contains 10 types of natural features (basin, bay, bend, crater, gap, gut, island, lake, ridge, and valley), and we selected five (bay, bend, crater, island, and lake) features and a total of 540 images for the experiments. Some features were not selected because they are uncommon (e.g., gap and gut), and some (e.g., basin, valley, and ridge) are extremely difficult to be visually inspected using optical remote sensing images alone due to their limited color representations. This five-category GeoNat dataset is trained using the same experimental setting as the first dataset. The saliency maps are generated on a trained VGG model with 98.89\% training accuracy and 82.52\% testing accuracy. The results reported in Appendix I show that despite small differences, the findings about the GeoAI models’ characteristics using the two datasets are quite consistent. This further supports the conclusion drawn in Table \ref{tbl_1}. 

\section{Conclusion}

As AI becomes an important tool for high-stakes decision-making, its explainability—the articulation of an AI’s algorithm’s rationale in deriving an answer—has become a critical research topic because it can help open the black box and help users gain the confidence and trust to adopt AI in real-world decision-making processes \citep{phillips2020four}. However, little research has assessed the capabilities of existing AI model explanation methods and their applicability in geospatial applications. This paper fills this knowledge gap by providing an in-depth analysis of existing methods’ mechanisms, especially those aiming to explain a deep learning model by capturing the “visual attention” of a machine in a saliency map. Multiple methods, including occlusion, Grad-CAM, and integrated gradients, were implemented and applied to a deep learning model for image classification tasks. By examining the decisions the model made for every single image of the training dataset (indicated in the saliency maps), we derived and summarized a number of model explanation goals (e.g., the ability to identify multiple objects of the same type) and assessed each method’s ability to meet such goals. The experiments used two natural feature datasets, and the results derived from these datasets are highly consistent, demonstrating good generalizability of the research findings in natural feature analysis. 

The results in Table \ref{tbl_1} show that no single AI explanation method can achieve all the explanation goals in a geospatial task. Therefore, simultaneously applying multiple methods when attempting to explain a GeoAI model is important so that the results can be cross-validated and uncertainties can be removed. The term 'uncertainty' is used to describe situations where a saliency map contains artifacts that do not reflect the model's learned knowledge but are instead caused by the inherent limitations of the model explanation methods (e.g., gradient saturation). Furthermore, while two datasets were used in the experiments to demonstrate the generalizability of our research findings, we cannot conclusively establish their universal generalizability without conducting more systematic experiments. This limitation underscores the need for further improvement in this area of research.

In the future, in addition to increasing the results accuracy of the GeoAI explanation methods, we will work to combine the global and per-decision model explanation algorithms and develop new strategies to support the automatic extraction of semantically meaningful concepts from saliency maps to further enhance the human understanding of machine learning processes. We will compare the consistency between human-understandable concepts defined in domain knowledge graphs (\citep{li2023geographvis, janowicz2022know, li2012semantic}) and the machine-learned features to further improve the explainability of the machine learning process. We will also use these methods to analyze the failure cases and understand why a GeoAI model makes a wrong decision. As discussed, detection of some natural features—especially those existing in hilly terrains, such as ridges and valleys—is often difficult to achieve using satellite images alone. This task can benefit from the use of additional data sources, such as digital elevation models (DEMs) data \citep{wang2021geoai, li2022geoimagenet}. Hence, our future work will also include extending the application of deep learning model explanation algorithms to a multisource learning framework. Finally, more experiments will be conducted to further verify the generalizability of our research findings using additional, more diverse datasets, including both natural and artificial features.
 
\section*{Acknowledgement}
This work is supported in part by the National Science Foundation under awards 2120943, 2230034, 1853864. 

\section*{Data and codes availability statement}
The data and codes that support the findings of this study are available at \url{https://github.com/ASUcicilab/explainable-geoai}. Instructions on how to use the data and codes are provided in the README file. 

\section*{Notes on contributors}
Chia-Yu Hsu is a research professional at Arizona State University. His research interests include artificial intelligence, computer vision, spatiotemporal data analysis, and their applications in climate change and terrain research.

Wenwen Li is a professor in geographic information science at Arizona State University (ASU). Her research interests are cyberinfrastructure, big data, GeoAI and their applications in data- and computation-intensive environmental and social sciences. At ASU, she directs the Cyberinfrastructure and Computational Intelligence Lab (http://cici.lab.asu.edu/) and serves as the Research Director for the Spatial Analysis Research Center. 

\bibliographystyle{tfv}
\bibliography{reference}

\end{document}

\begin{table}
\tbl{Statistics on percentage of saliency maps generated by different methods that can meet an explanation goal. ``Total'' means number of images in the GeoNat v1.0 dateset (Arundel et al. 2020) that supports the corresponding criterion listed in each row. }
{\begin{tabular}{p{0.4\textwidth}cccc} \toprule
 Model Explanation Goals & Occlusion Approach  & Grad-CAM  & Integrated Gradients  & Total \\ \midrule
 Ability to identify multiple objects of the same type & 29\% & 74\% & 81\% & 119 \\
 Ability to identify multiple prominent features of a single object & 32\% & 86\% & 94\% & 142 \\
 Ability to accurately detect object shape & 78\% & 48\% & 81\% & 540 \\
 Ability to correctly highlight important image regions	& 84\% & 60\% & 89\% & 540 \\
 Ability to generate clear attentions for an image with multiple high-contrast areas & 89\% & 82\% & 52\% & 132 \\
 Ability to generate clear attentions for an image when the contrast between the target and the background is low & 84\% & 81\% & 52\% & 204 \\
 \bottomrule
\end{tabular}}
\label{tbl_3}
\end{table}